\documentclass{article}

\PassOptionsToPackage{numbers, sort&compress}{natbib}
\usepackage[final]{neurips_2024}

\usepackage[toc,page,header]{appendix}
\usepackage{minitoc}

\usepackage[utf8]{inputenc} 
\usepackage[T1]{fontenc}    
\usepackage{hyperref}       
\usepackage{url}            
\usepackage{booktabs}       
\usepackage{amsfonts}       
\usepackage{nicefrac}       
\usepackage{microtype}      
\usepackage{xcolor}         
\usepackage{enumitem}
\usepackage{wrapfig}
\usepackage{microtype}
\usepackage{graphicx}
\usepackage{arydshln}
\usepackage{booktabs} 
\usepackage{tabularx}
\usepackage{ltablex}
\usepackage{makecell}
\usepackage{algorithm}

\usepackage{hyperref}
\usepackage{makecell}
\usepackage{multirow}
\usepackage{subcaption}

\usepackage{amsmath}
\usepackage{amssymb}
\usepackage{mathtools}
\usepackage{amsthm}
\newtheorem{definition}{Definition}

\usepackage{hyperref}
\hypersetup{
    colorlinks = true,
    linkcolor = blue,
    anchorcolor = blue,
    citecolor = black,
    filecolor = blue,
    urlcolor = blue
    }

\definecolor{headercolor}{rgb}{0.42, 0.56, 0.14} 
\definecolor{rowcolor}{rgb}{0.9, 0.9, 0.9} 
\definecolor{rebeccapurple}{HTML}{663399} 
\usepackage{colortbl}
\definecolor{Gray}{gray}{0.9}
\newcolumntype{g}{>{\columncolor{Gray}}c}

\usepackage[most]{tcolorbox}  
\usepackage{listings}

\newtcolorbox[auto counter, number within=section]{promptbox}[2][]{ 
  colback=gray!5!white,    
  colframe=blue!75!black,  
  coltitle=white,          
  title=#2, 
  sharp corners,           
  boxrule=0.5pt,           
  boxsep=5pt,              
  left=-1pt,                
  right=1pt,               
  top=5pt,                 
  bottom=5pt,              
  #1                        
}

\newtcolorbox[auto counter, number within=section]{evalbox}[2][]{ 
  colback=gray!5!white,    
  colframe=green!75!black,  
  coltitle=white,          
  title=#2, 
  sharp corners,           
  boxrule=0.5pt,           
  boxsep=5pt,              
  left=-1pt,                
  right=1pt,               
  top=5pt,                 
  bottom=5pt,              
  #1                        
}

\newtcolorbox{shadebox}[1][]{%
    colback=gray!10,
    colframe=white, 
    boxsep=5pt,
    arc=0mm,
    top=2pt,
    bottom=2pt,
    breakable,
    left=2pt,
    right=3pt,
    auto outer arc,
    #1 
}

\usepackage{amsmath}
\usepackage{amssymb}
\usepackage{mathtools}
\usepackage{amsthm}
\usepackage{tablefootnote}

\usepackage{algorithm}
\usepackage{algpseudocode}
\usepackage{adjustbox}

\usepackage{hyperref}
\usepackage{multirow}
\usepackage{graphicx}

\usepackage{pifont}
\definecolor{ForestGreen}{RGB}{34, 139, 34}

\usepackage{pgfplots}
\usepackage[framemethod=TikZ]{mdframed}
\usepackage{comment}
\usepackage{fontawesome}
\usepackage{pgfplotstable}
\usepackage{colortbl}

\definecolor{customgreen}{RGB}{116, 154, 114}
\definecolor{lightgreen}{RGB}{240, 246, 232}
\definecolor{greylight}{RGB}{242, 242, 242}
\definecolor{greydark}{RGB}{179, 179, 179}
\definecolor{ForestGreen}{RGB}{34, 139, 34}

\mdfsetup{%
    middlelinecolor =   none,
    middlelinewidth =   1pt,
    backgroundcolor =   blue!5,
    roundcorner     =   5pt,
}

\usepackage{tikz}
\usetikzlibrary{positioning, arrows.meta, shapes}

\newcommand{\lightbulbicon}{%
  \begin{tikzpicture}[baseline=-0.5ex]
    \draw[fill=white, draw=customgreen, thick] (0,0) circle (1.5ex);
    \node[scale=0.8, color=customgreen] at (0,0) {\faLightbulbO~};
  \end{tikzpicture}%
}

\newmdenv[
  rightline=false,
  linecolor=customgreen,
  outerlinewidth=2pt,
  topline=false,
  bottomline=false,
  leftline=true,
  skipabove=0.5\baselineskip,
  skipbelow=0.5\baselineskip,
  backgroundcolor=lightgreen,
  innerleftmargin=5pt, 
  innerrightmargin=5pt,
  innertopmargin=5pt,
  innerbottommargin=3pt,
  font=\fontfamily{lmr}\selectfont,
  roundcorner=5pt,
  singleextra={
    \node[xshift=-1.5pt] at (P-|O) {\lightbulbicon};
  },
  leftmargin=-17pt, 
  rightmargin=8pt,
]{customblockquote}

\title{Matchmaker: Self-Improving Large Language Model Programs for Schema Matching}

\author{%
  Nabeel Seedat \\
  University of Cambridge\\
  \texttt{ns741@cam.ac.uk} 
  \And
  Mihaela van der Schaar \\
  University of Cambridge\\
  \texttt{mv472@cam.ac.uk}
}

\begin{document}

\doparttoc
\faketableofcontents

\maketitle

\begin{abstract}
Schema matching -- the task of finding matches between attributes across disparate data sources with different tables and hierarchies -- is critical for creating interoperable machine learning (ML)-ready data. Addressing this fundamental data-centric problem has wide implications, especially in domains like healthcare, finance and e-commerce --- but also has the potential to benefit ML models more generally, by increasing the data available for ML model training. However, schema matching is a challenging ML task due to structural/hierarchical and semantic heterogeneity between different schemas. Previous ML approaches to automate schema matching have either required significant labeled data for model training, which is often unrealistic, or suffer from poor zero-shot performance. To this end, we propose Matchmaker -  a compositional language model program for schema matching, comprised of candidate generation, refinement and confidence scoring. Matchmaker also self-improves in a zero-shot manner without the need for labeled demonstrations via a novel optimization approach, which constructs synthetic in-context demonstrations to guide the language model's reasoning process.  Empirically, we demonstrate on real-world medical schema matching benchmarks that Matchmaker outperforms previous ML-based approaches, highlighting its potential to accelerate data integration and interoperability of ML-ready data.
\end{abstract}

\section{Introduction}\label{sec:intro}

Data is fundamental to the success of machine learning (ML) models, which depend on access to large, integrated and interoperable datasets \cite{jain2020overview,gupta2021dataB,renggli2021data,Sambasivan}. Although well-structured and uniform datasets like those on Kaggle are commonly assumed as the norm, such data is a rare luxury in practice. In real-world scenarios, tabular data often exists in heterogeneous and disparate databases with diverse formats, schemas, and terminologies, requiring harmonization to make the data "ML-ready" and interoperable. The heterogeneity of databases presents three critical issues for ML: (1) data harmonization and integration is an arduous task. Hence, researchers often limit the features/covariates used for model training to a smaller, often common, set of features \cite{avati2021beds,si2021deep,rajkomar2018scalable}, thereby limiting the potential performance of their ML models; (2) even if all the features are used, the lack of data interoperability means limited external validation of ML models \cite{balch2023machine, lehne2019digital, williams2022using, tiwari2020assessment,colubri2019machine}, which can undermine the credibility and utility of the ML models; and (3) missed opportunities for insights on larger harmonized datasets (e.g., larger patient populations), which may not be apparent when analyzing data sources independently.

\begin{figure}[!t]
\vspace{-1mm}
    \centering
\includegraphics[width=0.95\linewidth]{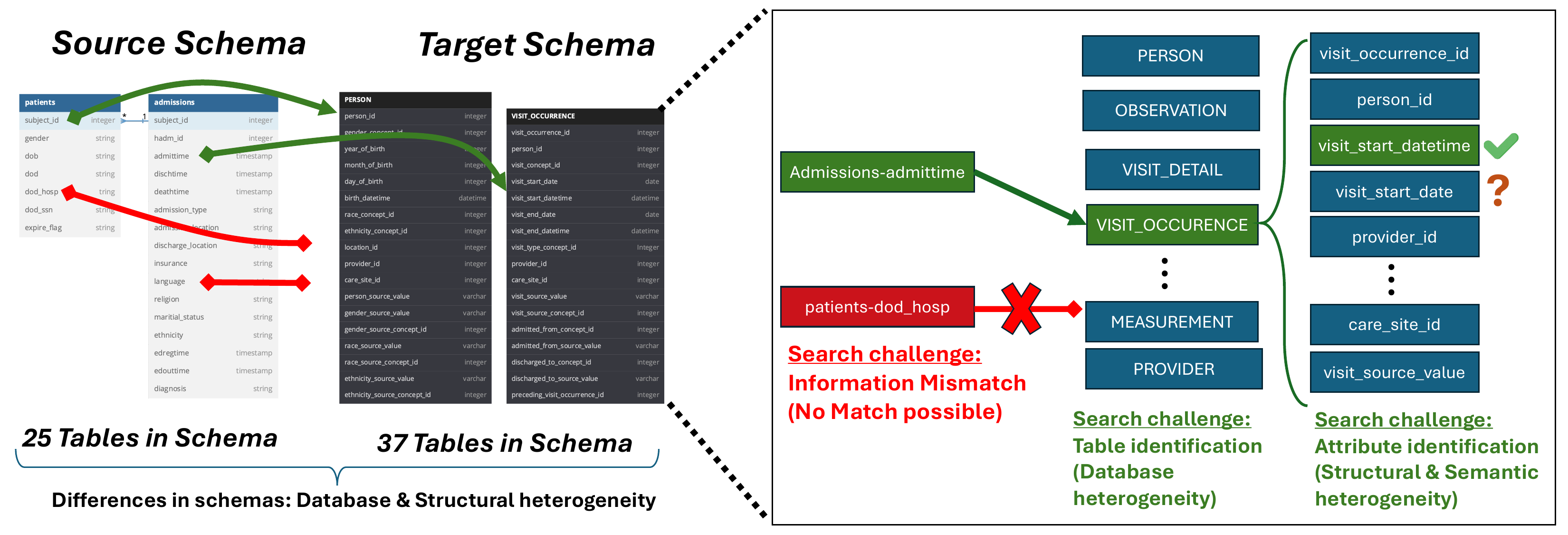}
\vspace{-0mm}
    \caption{\footnotesize     
    Example showing the complexity of schema matching due to the multi-faceted challenges:
\textcolor{ForestGreen}{\footnotesize \textbf{Database heterogeneity (green arrows):}} Identifying the correct target table is the first step, as each schema has a different number of tables, the corresponding information may be distributed differently across tables in each schema.
\textcolor{ForestGreen}{\textbf{Structural heterogeneity (green arrows):}} Once the appropriate table is found, matching attributes is complicated by differences in schema architectures, hierarchies, and granularity.
\textcolor{ForestGreen}{\textbf{Textual heterogeneity (green arrows):}} Ambiguity in matching when attributes have the same names but different meanings, or different names with the same meaning.
\textcolor{red}{\textbf{Information mismatch (red arrows):}} Some attributes in one schema may lack a corresponding match in the other schema, adding to the complexity of the matching process.}
    \label{fig:problem}
    \vspace{-2mm}
    \rule{\linewidth}{.5pt}
\vspace{-7mm}
\end{figure}

Schema matching is a critical first step in data harmonization, aiming to establish correspondences between attributes (i.e., features/covariates) measured across different data sources. Once matched, these correspondences can help harmonize data from disparate sources into a cohesive, ML-ready format. To understand the concept of schema matching, let us unpack the components of a schema.  A schema defines how data is organized in a database, comprising different tables (collections of related data entries) and columns (also known as "attributes" or "features") that represent specific data fields. Importantly, schemas go beyond simple tabular data commonly found in CSV files, as they capture the hierarchical structure and relationships between different tables and their attributes. For example, in healthcare, schemas from different hospitals may have varying tables and attributes representing patient information, lab measurements, diagnoses and treatments, with complex relationships and hierarchies connecting the tables.  Consequently, 
schema matching involves analyzing the context of attributes within the schema hierarchy to establish meaningful mappings that preserve the intended semantics and relationships. It goes beyond simple one-to-one column matching, considering not only the attribute itself but also the hierarchical structure and relationships between tables defined by the schema. Notably, schema matching does not assume access to raw data, relying only attribute names, descriptions and metadata (e.g., in healthcare, patient data cannot be queried or accessed directly due to privacy concerns or regulations \cite{Zhang2021SMATAA}).

The importance and value of schema matching cannot be overstated, as integrating data from various data sources such as different regions, organizations or applications is vital in healthcare but also in finance and e-commerce \cite{sheetrit2024rematch, Zhang2021SMATAA,el2024overview}. Schema matching is also generally valuable to \emph{anyone} working on ML, as a step toward increasing the training and validation data available to the ML community. e.g, in healthcare, integrating data from multiple hospitals can lead to more comprehensive datasets to train more generalizable ML prognostic models \citep{goetz2024generalization}. Similarly, in e-commerce, combining diverse customer data from various platforms can enable more accurate ML models built on customer data.

Unfortunately, prior ML approaches for "automated" schema matching often require extensive labeled data to train models \cite{Li2020DeepEM, Zhang2021SMATAA}, which is often infeasible. Although LLM-based methods \cite{Narayan2022CanFM, Mirchandani2023LargeLM} have attempted to address this, they have poor zero-shot performance and poor scalability in terms of the number of LLM calls. These limitations have hindered the adoption of ML for schema matching, meaning schema matching is still a largely manual and time-consuming task. To highlight the need for automated and better performing ML schema matching, in the healthcare domain, it took 500 hours for two experts to map the schemas between the MIMIC database and the OMOP common data model \cite{Paris2021TransformationAE}, demonstrating the substantial and non-trivial effort required.

Despite the need, schema matching is a challenging ML task, as shown in Fig. \ref{fig:problem}, as without access to the raw data, schema matching methods must rely only on the attribute names and other metadata to infer correspondences between attributes across schemas. This requires reasoning about various challenges, namely: $\blacktriangleright$ \textbf{Semantic heterogeneity:} ambiguous potential mappings, where attributes across schemas might have the same name but different meanings, or different names but the same meaning. $\blacktriangleright$ \textbf{Structural heterogeneity:} schemas that have varied architectures, hierarchies, and representational granularity. 
$\blacktriangleright$ \textbf{Database heterogeneity:} schemas having different numbers of tables in which information is represented. e.g. source schema table information may be represented across multiple target schema tables. Hence, it is non-trivial to identify the appropriate table for an attribute. 
$\blacktriangleright$ \textbf{Information mismatch:} Information may be contained in one schema, but not in another schema. Hence, reasoning about "no possible match" is as important as reasoning about a possible match.

\begin{wrapfigure}{r}{0.33\textwidth}
\vspace{-5mm}
  \centering
    \includegraphics[width=0.33\textwidth]{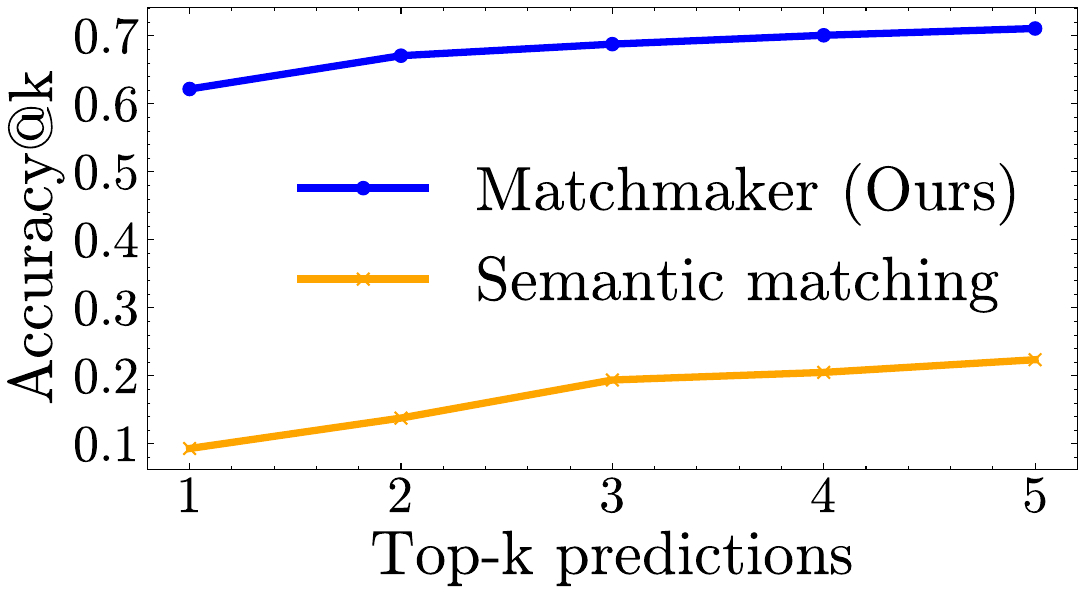}
    \vspace{-6mm}
   \caption{\footnotesize{Example result shows semantic similarity alone cannot solve schema matching, with low accuracy@k, compared to Matchmaker.}}
   \vspace{-2mm}
    \rule{\linewidth}{.5pt}
    \label{fig:motivation}
    \vspace{-8mm}
\end{wrapfigure}

These issues make schema matching a challenging task that cannot be solved by simple methods such as semantic similarity alone (see Fig. \ref{fig:motivation}). 
To this end, we introduce \emph{Matchmaker}, a self-improving compositional language model program for schema matching. Matchmaker leverages the reasoning capabilities of large language models (LLMs) via a compositional language model program with multi-stage LLM calls that comprise candidate generation, refinement, and confidence scoring (see Appendix \ref{appendix:examples} for examples of this process). Matchmaker also \emph{self-improves} without labeled data, via a novel optimization process using \emph{synthetic in-context examples} for the different stages of the language model program. Matchmaker makes the following contributions:

\begin{mdframed}[leftmargin=0pt, rightmargin=0pt, innerleftmargin=1pt, innerrightmargin=1pt, skipbelow=0pt]
\textbf{Contributions:} \textbf{\textcolor{ForestGreen}{\textcircled{1}}} We address recent calls to develop ML methods for data harmonization/interoperability \cite{balagopalan2024machine,gilbert2024augmented}.
\textbf{\textcolor{ForestGreen}{\textcircled{2}}} We propose Matchmaker, a compositional language model program to address the complexities of schema matching.
\textbf{\textcolor{ForestGreen}{\textcircled{3}}}  We introduce a novel optimization mechanism allowing Matchmaker to self-improve in a zero-shot manner via synthetic in-context examples that guide Matchmaker's reasoning process.
\textbf{\textcolor{ForestGreen}{\textcircled{4}}} We empirically demonstrate that Matchmaker outperforms different ML baselines on real-world schema matching benchmarks, along with showing the value of our self-improvement mechanism and how Matchmaker can be used with a human-in-the-loop. 
\end{mdframed}

\vspace{-5mm}
\section{Related Work}
\label{sec:related_work}
\vspace{-1mm}

This work engages with literature on schema matching (see Fig. \ref{fig:solution}) and contributes to data-centric AI.

\textbf{Schema matching.}
Previous ML-based schema matching approaches have shown promise, but suffer from limitations that hinder their practical applicability. Early works ~\cite{Mudgal2018DeepLF,Shraga2020ADnEVCS, Li2020DeepEM} computed similarity scores between schemas \cite{Do2002COMAA,Gal2011UncertainSM},
but focused on the simpler entity matching task (matching items within columns) rather than the more complex schema matching problem. Recent approaches like SMAT~\cite{Zhang2021SMATAA}, address full schema matching via deep learning (i.e. attention), but require substantial labeled matches for model training (> 50\%), making it impractical for real-world settings where labeled data is scarce or expensive to obtain (e.g. requiring experts).

To reduce the need for labels, LLMs have been applied to schema matching ~\cite{Zhang2023LargeDP, Narayan2022CanFM, zhang2023jellyfish}. Unfortunately, methods like LLM-DP using pre-trained LLMs \cite{Zhang2023LargeDP, Narayan2022CanFM} or Jellyfish fine-tuning LLMs \cite{zhang2023jellyfish} have been shown to have poor zero-shot performance (see Sec.~\ref{sec:exps}). Performance improvements were obtained with human-labeled examples of $\pm$500 examples, from which in-context examples are selected. However, reliance on human labeling is often unrealistic, limiting applicability. Additionally, LLM methods, like deep learning ones (e.g. SMAT \cite{Zhang2021SMATAA}), formulate schema matching as a binary classification task over the full Cartesian product of source and target schema attributes. For each pair of source-target attributes, the LLM is prompted to provide a label of Yes/No for the match (i.e. Is attribute A related to Attribute B? yes/no). The consequence is poor scalability ($O(n^2)$), which is computationally expensive for large schemas and costly due to the large number of LLM calls.

The closest work to ours is ReMatch \cite{sheetrit2024rematch}, which uses retrieval to find semantically similar candidate matches, thus reducing the search space. An LLM is then prompted to match a source schema attribute with retrieved target schema candidates. However, ReMatch relies solely on semantic matching, which we empirically demonstrate in Sec. \ref{sec:exps} does not suffice for real-world schemas. Our approach Matchmaker diverges from ReMatch along three dimensions: (1) \textit{System}:  ReMatch uses a single LLM call, while Matchmaker decomposes the task into a multi-stage compositional LLM program with multiple reasoning steps. 
(2) \textit{Candidate generation}: ReMatch only generates candidates via semantic retrieval, while Matchmaker incorporates \emph{multiple} candidate generation sources, including retrieval for semantic candidates and an LLM for contextual reasoning candidates. (3) \textit{Optimization}: ReMatch has a fixed LLM prompt template, while Matchmaker is an LLM program where we optimize the prompts via synthetic in-context examples.

\textbf{Data-Centric AI.}
Data-centric AI aims to systematically improve data quality for ML \cite{zha2023data,whang2023data,seedat2023navigating} through methods such as sample selection \cite{seedat2023dissecting, seedat2024triage} and \cite{jiang2023opendataval} of pre-existing integrated datasets. This work addresses a fundamental upstream problem: schema matching which enables the creation of harmonized datasets. In doing so, it contributes to the data-centric AI literature by tackling a critical issue that precedes and supports existing approaches to enhance data quality for ML.

\vspace{-4mm}
\section{Schema Matching}

\vspace{-2mm}
\subsection{Preliminaries.} 
\vspace{-1mm}
Consider the schema matching task, where the goal is to map attributes from a source schema ($S_s$) to a target schema ($S_t$). Each schema $S$ is defined as a collection of tables $\mathcal{T} = \{T_1, T_2, \ldots, T_m\}$. Each table $T_i$ contains a set of attributes $\mathcal{A}_i = \{A_{i1}, A_{i2}, \ldots, A_{ik}\}$. Additionally, each table $T_i$ is associated with metadata $m_i$ describing the purpose and content of the table. Similarly, each attribute $A_{ij}$ is associated with a description $d_{ij}$, which includes information describing the attribute, its data type and relational context. These descriptions and data types provide additional contextual information about the attributes to aid in the matching process. 

The schema matching task, defined below, aims to find matches between attributes across different schemas, respecting the database hierarchies, relationships and restrictions. Recall that schema matching operates solely on schema-level information (attributes and metadata), without having access to the raw data. This adds to the complexity, as matching must be performed without the benefit of analyzing the actual data values.

\begin{definition}[Schema Matching]
The goal of schema matching is to find a mapping function $f: \mathcal{A}_s \rightarrow \mathcal{A}_t \cup \{\varnothing\}$ that correctly assigns each attribute of the source schema $S_s$ to a corresponding attribute in the target schema $S_t$ or to the empty set $\varnothing$, indicating no possible match. 
\end{definition}

\vspace{-2mm}
\subsection{Schema matching as information retrieval.} 
As outlined in Sec.~\ref{sec:related_work}, schema matching is often formulated as a supervised binary classification problem (match/no match) over the entire Cartesian product of source and target schema attributes.
Beyond the computational side, this formulation has several drawbacks:
$\blacktriangleright$ \textbf{Labeling Cost:} It necessitates manual annotation of attribute pairs by domain experts, which is time-consuming and costly.
$\blacktriangleright$  \textbf{Class Imbalance:} The prevalence of non-matching attribute pairs significantly outnumbers matching pairs, resulting in severe class imbalance.
$\blacktriangleright$  \textbf{Lack of Ranking:} It does not yield a ranked list of candidate matches, which is critical for human review if multiple possible matches exist.

To address the drawbacks, we propose a two-stage information retrieval approach to schema matching:

$\blacktriangleright$ \textbf{1. Candidate generation}: For each source query attribute $A_{si} \in \mathcal{A}_s$ from the source schema $S_s$, we generate a set of potential matches from the target schema. Let $C_i \subseteq \mathcal{A}_t$ be the set of candidate target matches for query attribute $A_{si}$. The candidate generation process can be defined as a function $g: \mathcal{A}_s \times \mathcal{A}_t \rightarrow \mathcal{P}(\mathcal{A}_t)$, where $\mathcal{P}(\mathcal{A}_t)$ denotes the power set of $\mathcal{A}_t$, such that $C_i = g(A_{si}, \mathcal{A}_t)$.

$\blacktriangleright$ \textbf{2. Ranking}: We rank the candidates based on their relevance to the query attribute. We define a ranking function $r: (\mathcal{A}_s \times \mathcal{D}_s) \times (\mathcal{A}_t \times \mathcal{D}_t) \rightarrow \mathbb{R}$, where $\mathcal{D}_s$ and $\mathcal{D}_t$ represent the sets of contextual information associated with attributes in $\mathcal{A}_s$ and $\mathcal{A}_t$, respectively. 
For each source attribute $A_{si} \in \mathcal{A}_s$ and its associated contextual information $d_{si} \in \mathcal{D}_s$, the ranking function $r$ assigns a relevance score to each candidate attribute $A_{tj} \in C_i \subseteq \mathcal{A}_t$ and its associated contextual information $d_{tj} \in \mathcal{D}_t$:
$$r((A_{si}, d_{si}), (A_{tj}, d_{tj})) > r((A_{si}, d_{si}), (A_{tk}, d_{tk})) \Leftrightarrow A_{tj} \text{ is more relevant to } A_{si} \text{ than } A_{tk}.$$

The mapping function $f$ can then be defined as follows:
$$
f(A_{si}) = \begin{cases}
\arg\max_{A_{tj} \in C_i} r((A_{si}, d_{si}), (A_{tj}, d_{tj})), & \text{if } \max_{A_{tj} \in C_i} {r((A_{si}, d_{si}), (A_{tj}, d_{tj}))} \geq \tau \\
\varnothing, & \text{otherwise}
\end{cases}
$$
where $\tau$ is a relevance threshold and $f$ assigns the query attribute $A_{si}$ to the candidate attribute $A_{tj}$ with the highest relevance score. Conversely, we may assign $\varnothing$, indicating no match --- accounting for the fact that not all source attributes may have a possible match in the target schema.

\vspace{-1mm}
\section{Matchmaker: LLM-based Schema Matching}
\label{sec:method}
\vspace{-1mm}

We propose Matchmaker, a self-improving compositional language model (LM) program for schema matching (see Fig. \ref{fig:solution}), defined as a three-step LM program. For further details see Appendix \ref{appendix:algorithm}.

1.~\textbf{Multi-vector documents} (Sec. \ref{rag}): Creation of multi-vector documents from the target schema to facilitate semantic candidate retrieval of potential target attribute matches.\\
2.~\textbf{Candidate generation} (Sec. \ref{candidate-gen}): Employing two types of candidate generation: semantic retrieval and reasoning-based. The candidates are then refined into a smaller candidate set to evaluate.\\
3.~\textbf{Confidence scoring} (Sec.~\ref{confidence-scoring}): match confidence of a candidate target attribute to a query attribute.

\begin{mdframed}[leftmargin=0pt, rightmargin=0pt, innerleftmargin=1pt, innerrightmargin=1pt, skipbelow=0pt, backgroundcolor=ForestGreen!20]
\faLightbulbO~ \emph{Steps 1-3 define the unoptimized Matchmaker program. Finally, a key aspect of Matchmaker is our zero-shot optimization via synthetic in-context examples to improve performance (Sect. \ref{optimization}).}
\end{mdframed}
\vspace{-3mm}

\begin{figure}[!t]
    \vspace{-5mm}
    \centering
    \includegraphics[width=0.95\linewidth]{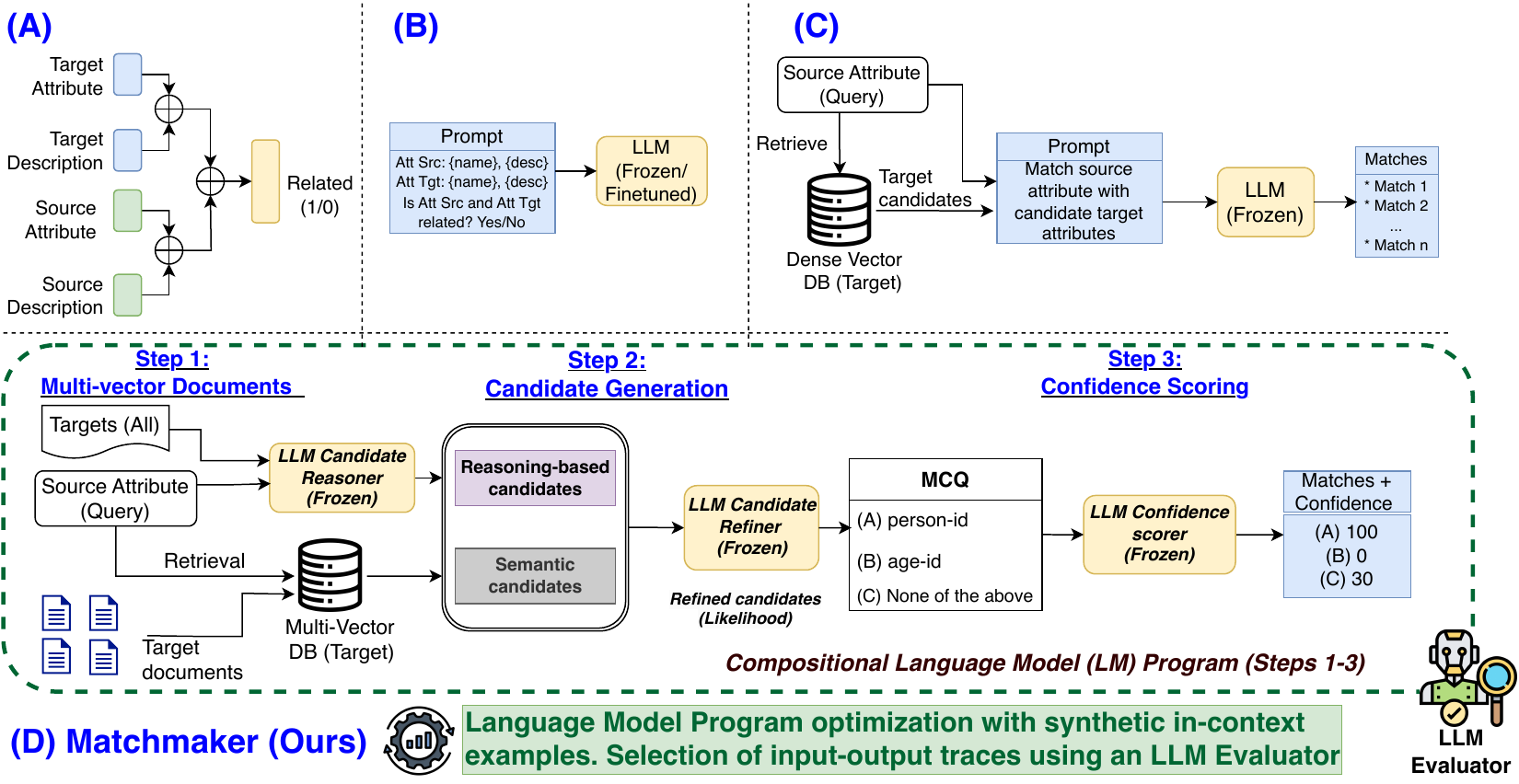}
    \vspace{-2mm}
    \caption{\footnotesize Conceptual comparison of different schema matching approaches. \textcolor{blue}{\textbf{(A)}} Supervised Matching \cite{Zhang2021SMATAA} employs a trained neural network (e.g., a transformer) to predict binary match/no-match labels across all attribute pairs, scaling as $\mathcal{O}(n)^2$ and requiring labeled data, thus unsuitable for zero-shot. 
\textcolor{blue}{\textbf{(B)}} LLM-Prompting \cite{Narayan2022CanFM,Zhang2023LargeDP} uses a frozen language model (e.g., GPT-4) for the same task, with similar scalability. Alternatively, \cite{zhang2023jellyfish} fine-tunes the LLM, which requires labeled data.
\textcolor{blue}{\textbf{(C)}} RAG-Based \cite{sheetrit2024rematch} improves scalability by retrieving candidates from a vector database and using a frozen LLM to select matches, but its effectiveness is limited to semantically similar options.
\textcolor{blue}{\textbf{(D)}} Matchmaker (Ours) performs schema matching via a self-improving, compositional language model program that enables enhanced reasoning. The program includes both retrieval and reasoning-based candidate generation with refinement and confidence scoring, allowing for better ranking. The program is optimized using synthetic in-context examples in the LLM prompts.}
 \vspace{-2mm}
\rule{\linewidth}{.5pt}
    \label{fig:solution}
    \vspace{-8mm}
\end{figure}

\textbf{Why LLMs for schema matching?}
Large Language Models (LLMs) form the foundation of Matchmaker, serving as key components within a compositional program comprised of multiple language model calls. Specifically, LLMs exhibit several appealing properties and capabilities for schema matching:
$\blacktriangleright$ \textbf{Contextual understanding}: LLMs have been pretrained on vast corpora of information, equipping them with extensive prior knowledge spanning different contexts and settings \cite{chowdhery2022palm, singhal2023large, cllm2024}. This contextual understanding enables LLMs to effectively reason about schema hierarchies and identify potential matches.
 $\blacktriangleright$ \textbf{Hypothesis proposers}: LLMs have been shown to be ``phenomenal hypothesis proposers'' \cite{qiu2023phenomenal}, making them particularly useful for candidate generation tasks.
$\blacktriangleright$ \textbf{Capable rankers}: LLMs have been shown to be highly capable at relevance ranking; assessing the suitability of candidates given a query and a set of options \cite{zhuang2023beyond,hou2024large}, especially ``when ranking candidates retrieved by multiple candidate generators'' \cite{hou2024large}.

\textbf{Defining a compositional LM program.} A compositional language model program, denoted as $\mathcal{L}$, is a multi-stage pipeline consisting of multiple LLM calls, i.e., $\mathcal{L} = \{l_1, l_2, \dots, l_n\}$, where $l_i: (s, k_s) \rightarrow \mathcal{Y}$ represents a specific LLM call taking as input a prompt string $s$ and in-context examples $k_s$ (which could be $\varnothing$). In the following sections (Secs. \ref{rag}-\ref{confidence-scoring}), we define the different components of $\mathcal{L}$ specific to Matchmaker. Finally, we describe our optimization process (Sec. \ref{optimization}).

\subsection{Multi-vector documents (Step 1)}\label{rag}
To facilitate efficient retrieval of semantically similar target schema candidates for any given source schema query, we construct a vector database containing target schema attributes. We begin by representing the target schema as a collection of structured documents. Specifically, for each table $T$ in the target schema $S_t$, we create a document consisting of the attribute names and append the attribute's textual description and data type, providing contextual information about each attribute. The metadata of each document includes the description of the table itself.

Unlike the common approach where each document is chunked and encoded as a single high-dimensional vector, Matchmaker employs multi-vector representations. Specifically, we use ColBERT-v2 \cite{santhanam2022colbertv2} model to encode the document chunks, producing an embedding per token (i.e., token-level dense vector), capturing token-level interactions. This approach has been demonstrated to enable better expressivity \cite{thakur2021beir, lee2024rethinking} and out-of-domain performance \cite{santhanam2022colbertv2}. In the next section, we detail how we retrieve semantically similar candidates for a given query using this multi-vector representation.

\vspace{-2mm}
\subsection{Diverse candidate generation (Step 2)}\label{candidate-gen}
\vspace{-1mm}
To narrow down the search space, Matchmaker identifies a subset of candidate attributes from the target schema that are likely matches for a query attribute $q_i \in A_s$ from the source schema. We draw inspiration from \cite{hou2024large}, which demonstrates that LLM ranking performance improves ```when ranking candidates are retrieved by multiple candidate generators.'' Hence, while semantic candidates are commonly used, Matchmaker goes beyond and employs two distinct types of candidate generation: (i) Semantic retrieval candidates retrieved from the vector database, and (ii) Reasoning-based candidates using a language model. This is then followed by a candidate refinement step. We outline each type of candidate generation applicable to a given query attribute $q_i \in A_s$.

\textbf{(i) Semantic retrieval candidates.} Given query $q_i$, we encode it using ColBERT-V2, obtaining a multi-vector query embedding.  Matchmaker then uses this query embedding to retrieve the top-k matching target schema attributes in the vector database. The top-k semantically similar candidates are denoted as $\mathcal{C}_s$.  We model similarity via late-interaction \cite{khattab2020colbert}, where each query embedding interacts with all document embeddings via a MaxSim operator, which computes the maximum similarity (e.g., cosine similarity), and finally the scalar outputs of each of these operators are summed across the different query terms. 

\textbf{(ii) Reasoning-based candidates.} To complement semantic matches, Matchmaker generates reasoning-based candidates using a candidate reasoner LLM denoted as $l_c: (q_i, \mathcal{A}_t) \rightarrow \mathcal{C}_R$, where $q_i$ is the i-th query, $\mathcal{A}_t$ is the set of all target attributes and $\mathcal{C}_R$ is a reasoning-based candidate set. Specifically, Matchmaker employs Chain of Thought (CoT) prompting \cite{wei2022chain} to reason about the target attributes $\mathcal{A}_t$ given the context of the schema hierarchy, descriptions and data types --- generating the most likely and relevant target schema candidate matches for each query $q_i$.

\textbf{Refinement.}
At this stage, the set of candidates is $\mathcal{C} = \mathcal{C}_R \cup \mathcal{C}_s$. Given the diverse set of candidates, Matchmaker aims to determine which candidates are the most likely and relevant matches for a given query, to obtain a smaller candidate set $\mathcal{C}^*$ to score and rank. Candidate refinement is achieved with a refiner LLM using CoT, denoted as $l_r: s \rightarrow \mathcal{C}^*$, where $s = (\mathcal{C}, q_i)$ and $q_i$ is the i-th source query.

\vspace{-3mm}
\subsection{Confidence scoring (Step 3)}\label{confidence-scoring}
\vspace{-1mm}

The refined set of candidates, $\mathcal{C}^*$ remains unordered. Hence, this step aims to obtain confidence scores to rank the candidates but also gauge the certainty of each match, recognizing that sometimes no suitable source-to-target attribute match exists, which requires the system to abstain from making a match.
While language models may not be well-calibrated at the sequence level, recent research has shown that they exhibit better calibration at the token level \cite{ren2023self}, a feature notably beneficial in multiple-choice question (MCQ) tasks \cite{kadavath2022language}. Leveraging this insight, Matchmaker structures the candidate scoring task as an MCQ format, labeling each candidate in $\mathcal{C}^*$ for query $q_i$ as options \textit{(A), (B), (C), etc.} Additionally, to account for the possibility that none of the target attribute candidates are a good match or there might be no possible match in the target schema, Matchmaker includes an abstain option by adding "NONE of the above" as a choice. This ensures that the LLM is not forced to select a candidate when there is no suitable match, aligning with the practices in \cite{ren2023self,ding2023knowledge}.

Matchmaker finally performs candidate ranking, where it is common  to
evaluate each candidate individually \cite{hu2024rankprompt, wang2023large,zheng2024judging}. Confidence scores are obtained by prompting the LLM to reason about the relevance of each candidate $c_i \in \mathcal{C}^*$ to the given query $q_i$. Furthermore, prior work has shown that LLMs can provide good uncertainty at token-level \cite{kadavath2022language} like in our MCQ, which is achievable via prompting \cite{tian2023just}. Consequently, Matchmaker elicits a confidence score by prompting the LLM to provide a value between 0 and 100, indicating the relevance of a match. These confidence scores are then used to either rerank the candidates or, if the highest score is assigned to "None of the above," return an empty list, suggesting that no suitable matches exist for the given query.

\vspace{-1mm}
\subsection{Self-improvment: Zero-shot optimization using synthetic in-context examples}\label{optimization}
\vspace{-1mm}
Matchmaker optimizes the language model program $\mathcal{L}$ by leveraging the few-shot learning capabilities of LLMs \cite{brown2020language,agarwal2024many,dong2022survey}. This is achieved by selecting input-output demonstrations (i.e. in-context examples). In Sec. \ref{sec:exps}, we contrast this with an alternative self-improvement method via self-reflection.

However, selecting in-context examples is non-trivial for schema matching for two reasons. 
(i) \textbf{Lack of labeled demonstrations:} 
We do not have access to labeled input-output demonstrations from which to select in-context examples. To overcome this challenge, we use the unlabeled schemas to create a "evaluation" set  $\mathcal{D}_{eval} = \{e_1, e_2, \ldots, e_m\}$, made up of different types of source queries. Specifically, we identify "easy queries" where the top-n (n=5) target schema semantic matches have a similarity score $>0.95$, and "challenging queries" with the lowest semantic matches. (ii) \textbf{Lack of an evaluator:} To assess Matchmaker's capabilities on the evaluation set and guide the optimization process, we need a validation metric. Since no validator is readily available, we propose to use an evaluator LLM, $\mathcal{E}: (e_i, \mathcal{L}(e_i)) \rightarrow \mathbb{R}$, that employs chain of thought \cite{wei2022chain} to score the relevance (from 0-5) of matches obtained from $\mathcal{L}$ when evaluated on examples from $\mathcal{D}_{eval}$.

\begin{wrapfigure}{r}{0.475\linewidth}
\centering
\vspace{-6mm}
\scalebox{0.85}{%
\begin{minipage}{1\linewidth}
\vspace{-3mm}
\begin{algorithm}[H]
\caption{Optimize LM program $\mathcal{L}$}
\begin{algorithmic}[1]
\State \textbf{Input:} Set of evaluation queries $\mathcal{D}_{eval} = {e_1, e_2, \ldots, e_n}$
\State \textbf{Output:} Set of top $n$ demonstrations $D_{demo}$
\For{each input $e_i \in \mathcal{D}_{eval}$}
\State $\hat{y}_i, trace_i \leftarrow \mathcal{L}(e_i)$ \Comment{Teacher $\mathcal{L}$ predicts, storing outputs and intermediate traces}
\State $s_i \leftarrow \mathcal{E}(e_i, \hat{y}_i)$ \Comment{ Evaluation score }
\State $D_{demo} \leftarrow D_{demo} \cup {(e_i, trace_i, \hat{y}_i, s_i)}$
\EndFor
\State Sort $D_{demo}$ by score
\State \textbf{return} $D_{demo}[0:n]$ \Comment{Select top $n$}
\end{algorithmic}
\label{algorithm}
\end{algorithm}
\end{minipage}}
\vspace{-3mm}
\end{wrapfigure}

\textbf{Zero-shot optimization with synthetic in-context examples.} To optimize our multi-stage language model program, we aim to select in-context examples for each component in $\mathcal{L}$. However, in-context demonstrations for the intermediate stages are typically unavailable. 
To address this, we simulate \textit{traces} by running $\mathcal{L}$ on the evaluation examples $e_i \in \mathcal{D}_{eval}$.  A trace captures the intermediate input-output pairs of each component in $\mathcal{L}$ during the execution of $\mathcal{L}$ on a given example. We then score the final output using the evaluator $\mathcal{E}$, assessing the overall performance of $\mathcal{L}$ on that example.  We then adopt the DSPy bootstrapping process \cite{khattab2023dspy} that uses the intermediate input-output pairs from the traces that produced the highest evaluation scores as synthetic in-context examples for each component of $\mathcal{L}$. In other words, we use the input-output pairs generated by Matchmaker itself (which resulted in good evaluation performance) as synthetic in-context examples to guide the LLM reasoning.  This allows us to improve the program in a zero-shot manner, without relying on actual labeled data. Algorithm \ref{algorithm} provides an overview of the process. We refer to $\mathcal{L}$ with the selected in-context examples as Matchmaker (Optimized).

\vspace{-2mm}
\section{Experiments}\label{sec:exps}
\vspace{-2mm}

 We now empirically investigate multiple aspects of Matchmaker. For qualitative examples that illustrate Matchmaker's application, refer to Appendix \ref{appendix:examples}.

 \begin{table}[!h]
\vspace{-3mm}
\centering
\arrayrulecolor{rebeccapurple}
\setlength{\arrayrulewidth}{1pt}
\begin{adjustbox}{max width=0.99\textwidth}
\begin{tabular}{|p{0.6cm}|p{3.29cm}p{14.7cm}|}
\rowcolor[HTML]{663399} 
{\color[HTML]{FFFFFF} \textbf{Sec.}} & {\color[HTML]{FFFFFF} \textbf{Experiment}} & {\color[HTML]{FFFFFF} \textbf{Goal}} \\
\ref{exp1} & Overall performance & \textit{Performance of Matchmaker vs schema matching benchmarks} \\
\ref{exp2} & Self-improvement & \textit{Performance of Matchmaker: optimized vs unoptimized vs alternative improvement via self-reflection}  \\
\ref{exp3} & Source of gain & \textit{Ablation to understand Matchmakers candidate generation}  \\
\ref{exp4} & Matchmaker in practice & \textit{Using Matchmaker with humans: uncertainty deferral and remedial action}  \\ 
\addlinespace[-0.5ex]
\bottomrule
\end{tabular}
\end{adjustbox}
\vspace{-2mm}
\end{table}

  \textbf{Setup.} We conduct experiments on the MIMIC-OMOP and Synthea-OMOP datasets, which are the standard benchmark datasets used in prior schema matching works \citep{sheetrit2024rematch, zhang2023jellyfish, Narayan2022CanFM, Zhang2023LargeDP, Zhang2021SMATAA}. These datasets are real-world healthcare schema matching datasets and have been widely adopted due to their complexity and their reflection of real-world schema matching challenges. Additionally, complex, real-world schema matching datasets are rare and difficult to obtain, as annotating them requires extensive domain expertise (e.g., 500 hours for MIMIC-OMOP), making them invaluable test beds for schema matching algorithms. An overview of the datasets is provided in Appendix \ref{appendix:details}, along with further experimental details.

\textbf{Metrics.} We evaluate schema matching performance using accuracy@k used in \cite{sheetrit2024rematch} and is commonly used in information retrieval. Besides, ReMatch the other baselines treat schema matching as a binary classification using F1-score as the metric. In our setting of m:1 matching (i.e. one match for each query), accuracy@1 is equivalent to F1-score, if the binary label is assigned via $argmax$. Hence, we report accuracy@1 for all other baselines for comparison to retrieval based approaches. Unless otherwise stated, metrics are averaged over 5 seeds (with standard deviation).

\vspace{-3mm}
\subsection{Schema Matching performance: Does it work?} \label{exp1}
\vspace{-1mm}

Matchmaker's performance is compared to diverse schema-matching baselines (refer to Sec.~\ref{sec:related_work}). These include (i) LLM-based methods such as ReMatch and LLM-DP, (ii) the state-of-the-art non-LLM supervised model, SMAT, and (iii) Jellyfish, an LLM specifically fine-tuned for data preprocessing tasks, including schema matching.While Jellyfish is fine-tuned using the same MIMIC and Synthea datasets, giving it an advantage, we include it as a baseline to highlight Matchmaker's zero-shot performance using a general-purpose LLM. This comparison spans general-purpose LLMs, traditional supervised approaches, and task-specific fine-tuned models. All LLM baselines use GPT-4 (0613) \citep{openai2023gpt} as the backbone for fair comparison to the original works, as well as, mitigating variability due to the LLM itself.  Other LLM backbone results are found in Appendix \ref{appendix:more-exps}.

\textbf{Matchmaker has the best overall performance.} 
Matchmaker consistently outperforms baselines, across all settings, as shown in Table \ref{tab:overall}. Importantly, we find the largest performance gains \emph{(+-20\%)} for accuracy@1. This is a desirable property, as it suggests a better ranking of matches. Moreover, a higher accuracy at low $k$ values enables the use of smaller prediction sets, reducing the human effort required to select the final best target attribute match for a given source attribute query.

\textbf{Formulation as information retrieval outperforms binary classification.} 
A key insight from our experiments is that information retrieval-based approaches (Matchmaker and ReMatch) perform substantially better for accuracy@1 compared to the other binary classification-based approaches, which evaluate the full Cartesian product of attributes. This performance gap can be attributed to the smaller search space of the information retrieval formulation. Notably, Matchmaker and ReMatch are evaluated on all mappings, including matches and nulls ("No possible match"), whereas binary classification methods consider a simpler problem by only evaluating true matches.

\begin{table}[!t]
    \vspace{-5mm}
    \centering
        \caption{\footnotesize Comparison of schema matching performance of different baselines.}
        \scalebox{0.75}{
    \begin{tabular}{c|c|c|ccccccc}
        & & Matchmaker & ReMatch  & JellyFish-13b & Jellyfish-7b  & LLM-DP  & SMAT (20-80)  & SMAT (50-50) \\
        \hline \hline
        \multirow{3}{*}{\rotatebox{90}{MIMIC}} & acc@1 &  \bf 62.20 $\pm$  2.40 & 42.50  & 15.36 $\pm$ 5.00 & 14.25 $\pm$ 3.00  & 29.59 $\pm$ 2.00 & 6.05 $\pm$ 5.00  & 10.85 $\pm$ 6.00 \\
        & acc@3 & \bf  68.80 $\pm$ 2.00 & 63.80  & N.A. & N.A. & N.A. & N.A. & N.A.  \\
        & acc@5 &  \bf  71.10 $\pm$ 2.00 & \bf  72.90  & N.A. & N.A. & N.A. & N.A. & N.A. \\
         \hline\hline
        \multirow{3}{*}{\rotatebox{90}{Synthea}} & acc@1 & \bf  70.20 $\pm$ 1.70  & 50.50  & 35.17 $\pm$ 3.90 & 31.52 $\pm$ 1.70 & 41.44 $\pm$ 5.40 & 36.23 $\pm$ 3.30 & 44.88 $\pm$ 2.60   \\
        & acc@3 & \bf  78.60 $\pm$ 2.50 & 58.10  & N.A. & N.A. & N.A. & N.A. & N.A.  \\
        & acc@5 & \bf  80.90 $\pm$ 1.10 & 74.30  &  N.A. & N.A. & N.A. & N.A.  & N.A. \\
    \end{tabular}}
    \label{tab:overall}
    \vspace{-4mm}
\end{table}

\subsection{Matchmaker self-improvement analysis} \label{exp2}

Matchmaker self-improves its language model program in a zero-shot manner (no labeled examples) via an optimization process using synthetic in-context examples (Sec.~\ref{optimization}). We evaluate the performance of Matchmaker (Optimized) to three alternatives to disentangle the value of our in-context example selection mechanism: (1) Matchmaker (Vanilla), which is the vanilla language model program without in-context examples, (2) Matchmaker (Random): random selection of in-context examples rather than our optimized/systematic selection of in-context examples and (3) Matchmaker (Self-Reflection), which employs a self-reflection or self-refinement mechanism \citep{pan2023automatically,madaan2024self} as an alternative self-improvement approach. i.e. the LLM iteratively self-corrects through feedback and has been used for various LLM tasks to improve performance.

The results in Table \ref{tab:reflection} illustrate the following: $\blacktriangleright$ Matchmaker (Optimized) achieves significant performance gains compared to Matchmaker (Vanilla), particularly at low $k$ values (+-5\% improvement for acc@1). This finding highlights the value of the synthetic in-context examples and the potential for zero-shot self-improvement, even in the absence of labeled data or well-defined evaluation metrics. 
$\blacktriangleright$ Matchmaker (Optimized) outperforms Matchmaker (Random),  confirming that our systematic selection of in-context samples is the key driver of performance gains, rather than the mere inclusion of \emph{any} in-context examples. 
$\blacktriangleright$ Matchmaker (Optimized) which uses an LLM evaluator to score demonstration examples directly, provides better performance gains compared to the self-reflection approach, where an LLM simply self-refines along the pipeline. This underscores the importance of input-output demonstrations for Matchmaker, especially considering the multi-stage nature of the program, where the outputs of earlier components affect later components. 

\begin{table}[!h]
    \centering
    \vspace{-1mm}
    \caption{\footnotesize Comparison of different Matchmaker self-improvement mechanisms, showing the value of our systematic selection of in-context samples vs random selection, vanilla or improvement via self-reflection.}
    \scalebox{0.9}{
    \begin{tabular}{c|c|cccc}
        & & \makecell{Matchmaker \\ (Optimized)} & \makecell{Matchmaker \\(Random)} & \makecell{Matchmaker \\(Vanilla)} & \makecell{Matchmaker \\(Self-reflection) }\\
        \hline \hline
        \multirow{3}{*}{\rotatebox{90}{MIMIC}} & acc@1 & \bf 62.20 $\pm$ 2.40 & 55.36 $\pm$ 2.15 & 57.90 $\pm$ 1.20 & 57.10 $\pm$ 0.60 \\
        & acc@3 & \bf 68.80 $\pm$ 2.00 & 62.74 $\pm$ 4.50 & 66.40 $\pm$ 0.60 & 66.60 $\pm$ 1.00 \\
        & acc@5 & \bf 71.10 $\pm$ 2.00 & 65.00 $\pm$ 6.42 & 70.20 $\pm$ 0.70 & 70.60 $\pm$ 0.50 \\
        \hline
        \multirow{3}{*}{\rotatebox{90}{Synthea}} & acc@1 & \bf 70.20 $\pm$ 1.70 & 67.76 $\pm$ 1.38 & 65.40 $\pm$ 0.90 & 67.80 $\pm$ 1.40 \\
        & acc@3 & \bf 78.60 $\pm$ 2.50 & 76.19 $\pm$ 5.28 & 78.20 $\pm$ 0.60 & 75.90 $\pm$ 0.70 \\
        & acc@5 & 80.90 $\pm$ 1.10 & 77.66 $\pm$ 5.07 & \bf 83.20 $\pm$ 1.10 & 81.10 $\pm$ 1.90 \\
    \end{tabular}}
   \label{tab:reflection}
    \vspace{-0mm}
\end{table}

\subsection{Source of gain ablation: Why does it work?} \label{exp3}
\vspace{-0mm}
Matchmaker's performance relies on the generated candidate matches. Given its strong performance compared to baselines, we investigate which candidate generation approach contributes most to Matchmaker's success. To disentangle the role of each candidate generation method, we assess Matchmaker with (1) reasoning-based candidates from the LLM only (\texttt{Matchmaker\_reasoning\_only}) and (2) semantic candidates via retrieval only (\texttt{Matchmaker\_semantic\_only}).

The results in Table \ref{tab:ablation} show that reasoning-based candidates outperform semantic retrieval-based candidates. This finding suggests that LLM reasoning over the database hierarchy and data types produces better candidates than semantic matches that do not consider hierarchical relationships. In some cases (e.g., Synthea acc@1), the inclusion of retrieval-based candidates harms performance. However, the overall results indicate that Matchmaker benefits from both candidate generation approaches, with reasoning-based candidates providing greater value. This highlights the value of candidate generation via diverse mechanisms.

\begin{table}[!h]
    \vspace{-3mm}
    \centering
        \caption{\footnotesize Understanding the impact of different candidate generation approaches on Matchmaker. }
        \scalebox{0.8}{
    \begin{tabular}{c|c|c|cc}
        & & Matchmaker & \texttt{Matchmaker\_reasoning\_only}   & \texttt{Matchmaker\_semantic\_only}  \\
        \hline \hline
        \multirow{3}{*}{\rotatebox{90}{MIMIC}} & acc@1 &   62.20 $\pm$ 2.50 & 61.60 $\pm$ 1.50 & 60.20 $\pm$ 2.20   \\    
        & acc@3 &   68.80 $\pm$ 2.00 & 68.70 $\pm$ 1.60 & 64.50 $\pm$ 2.80 \\
        & acc@5 &    71.10 $\pm$ 2.00 & 70.40 $\pm$ 1.00 &  67.10 $\pm$ 3.10    \\
         \hline
        \multirow{3}{*}{\rotatebox{90}{Synthea}} & acc@1 & 70.20 $\pm$ 1.70  & 73.00 $\pm$ 1.90 &  63.10 $\pm$ 0.70\\
        & acc@3 & 78.60 $\pm$ 2.50 & 78.50 $\pm$ 1.50 & 77.40 $\pm$ 0.90 \\
        & acc@5 & 80.90 $\pm$ 1.10 & 79.40 $\pm$ 0.30 & 80.20 $\pm$ 0.40 \\
    \end{tabular}}

    \label{tab:ablation}
     \vspace{-3mm}
\end{table}

\subsection{Matchmaker in practice: Human-in-the-loop deferral and remedial action.} \label{exp4}

How might we use Matchmaker in practice for schema matching? Let us examine two cases.

\vspace{-1mm}
\begin{figure}[!h]
    \centering
    \begin{subfigure}[b]{0.4\textwidth}
        \includegraphics[width=\textwidth]{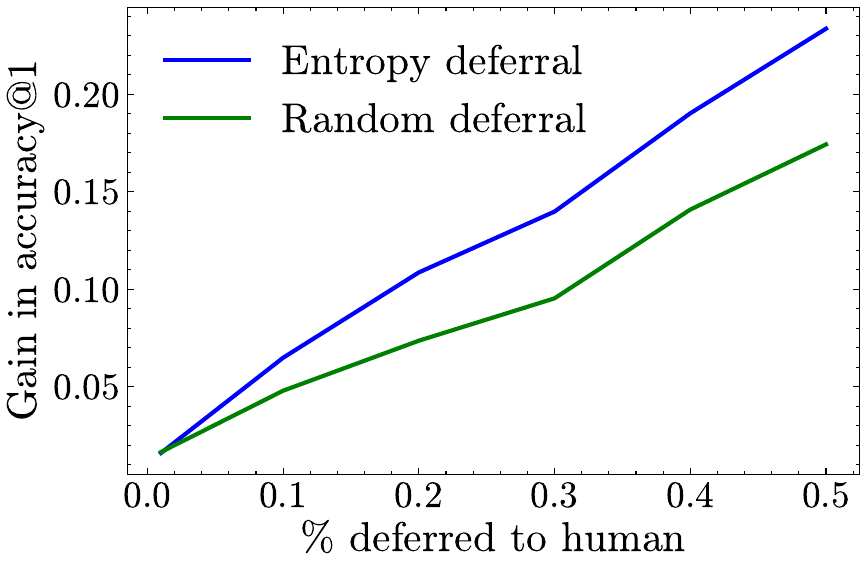}
         \vspace{-5mm}
        \caption{}
        \label{fig:sub1}
    \end{subfigure}
    \hspace{1cm}
    \begin{subfigure}[b]{0.4\textwidth}
        \includegraphics[width=\textwidth]{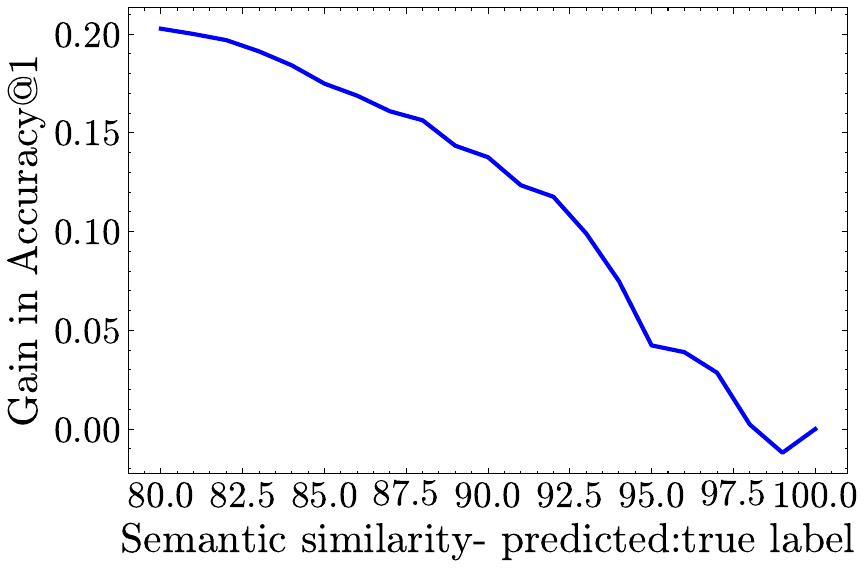}
         \vspace{-5mm}
        \caption{}
        \label{fig:sub2}
    \end{subfigure}
     \vspace{-2mm}
    \caption{\footnotesize Examples of using Matchmaker in practice. (a) Deferring uncertain samples to humans via entropy deferral improves schema matching performance. (b) Performance gains are obtained when correcting errors which are semantically similar to the true attribute. }
    \label{fig:errors}
     \vspace{-0mm}
\end{figure}

\textbf{(1) Matchmaker with human-in-the-loop deferral:}
We evaluate the effectiveness of integrating Matchmaker with a human-in-the-loop approach by deferring uncertain matches to human experts (i.e., an oracle) for correction.  High-uncertainty cases are identified using the entropy of Matchmaker's confidence scores, with the most challenging matches (those with the highest entropy) deferred to the oracle. We evaluate different deferral percentages $p \in \{0,10,20,30,40,50\}$ and observe that entropy-based deferral consistently yields greater performance gains compared to random deferral, as shown in Fig. \ref{fig:errors}(a). This finding highlights the practical value of Matchmaker in real-world settings, where based on entropy, one could strategically seek human oversight for challenging matches and improve overall schema matching performance. The appropriate deferral percentage, however, depends on context-specific factors such as human bandwidth and expert availability.

\textbf{(2) Evaluating ease of remedial action based on the similarity between incorrect predictions and true target attributes:}
Not all errors in source-target matching are equal; some might be easier to rectify than others. We hypothesize that errors involving semantically similar attributes are easier to correct compared to those involving completely dissimilar attributes. We analyze the cosine similarity between incorrectly predicted attributes and their true target attributes using Pubmed-Bert embeddings. To simulate post-hoc remedial action, we assess the performance gains achieved by correcting erroneous predictions that exceed different similarity thresholds. 
Figure \ref{fig:errors}(b) shows substantial improvements in accuracy@1 when "fixing" errors, with high semantic similarity between the erroneous prediction and true attribute (e.g., cosine similarity $\geq 0.8$). These results suggest that Matchmaker's incorrect predictions are often semantically close to the true attributes (i.e. our errors are not far off), making them more amenable to post-hoc remedial actions. This demonstrates the viability of post-hoc remedial actions to improve schema matching performance.

\section{Discussion}
Matchmaker introduces a novel approach to schema matching, using a self-improving compositional program using LLMs.  Matchmaker's superior performance compared to existing ML-based approaches, underlines its potential to accelerate data integration for ML-ready data. Matchmaker's zero-shot self-improvement mechanism, using synthetic in-context examples, showcases the potential of using LLMs to handle complex reasoning tasks without relying on labeled data.

\textbf{Limitations and opportunities.}
(1) Matchmaker, while effective in schema matching, represents just one component of the broader data harmonization process and needs to be integrated with other tasks to generate ML-ready data.
(2) Despite its advantages over alternative ML-based approaches, Matchmaker is not a panacea and does not achieve perfect automation. It is best used with a human-in-the-loop (Sec.~\ref{exp4}) to ensure reliability in real-world settings.

\section*{Acknowledgements}
NS is supported by the Cystic Fibrosis Trust. The authors thank the anonymous reviewers, Fergus Imrie, Nicolas Astorga, Julianna Piskorz and Andrew Rashbass for their feedback. The authors are grateful for the support of Microsoft’s Accelerate Foundation Models Academic Research initiative.

\clearpage
\bibliography{refs}
\bibliographystyle{unsrtnat}

\clearpage
\newpage
\appendix

\part{Appendix - Matchmaker: Self-Improving Large Language Model Programs for Schema Matching}
\mtcsetdepth{parttoc}{3} 
\parttoc
\newpage

\section{Matchmaker additional details}

\subsection{Matchmaker within the context of LLM table reasoning.}

There has recently been works on LLMs for table reasoning. We contrast them to Matchmaker along a variety of dimensions below.

\textbf{Task/Goal:}
The table reasoning papers tackle a variety of tasks centered around understanding and interacting with tabular data. Some examples include: TabSQLify \citep{nahid2024tabsqlify} and OPENTAB \citep{kong2023opentab} focus on table question answering and fact verification, aiming to extract relevant information from tables to answer questions or verify statements. Chain-of-Table \citep{wang2023chaintable} and "Large Language Models are Few-Shot Table Reasoners" \citep{chen2023fewshot} explore LLMs' capabilities in reasoning over tables for question answering and fact verification tasks. The survey paper "Large Language Model for Table Processing" \citep{lu2024large} covers a broader range of tasks, including table manipulation, table augmentation, and text-to-SQL conversion, showcasing LLMs' potential in interpreting and manipulating tabular data. In contrast, Matchmaker addresses the task of schema matching, which aims to find correspondences between attributes across different schemas or tables. The goal is to enable data integration by mapping attributes from a source schema to a target schema, considering the structural and semantic differences between them. This task is crucial for creating ML-ready datasets by harmonizing data from diverse sources.

\textbf{Approach:}
Table reasoning approaches span prompting LLMs for direct answers \citep{chen2023fewshot}, program synthesis to generate SQL/code \citep{nahid2024tabsqlify,kong2023opentab}, iterative table transformation \citep{wang2023chaintable}, instruction tuning \citep{lu2024large}, and agent-based methods \citep{lu2024large}. Matchmaker proposes a novel self-improving compositional language model program. It leverages LLM reasoning via a pipeline with multiple LLM calls for candidate generation, refinement and confidence scoring. It also self-improves without labeled data via synthetic in-context examples.

\textbf{Inputs:}
The table reasoning papers mostly focus on single tables as input along with a question/query.
Matchmaker takes as input two tables/schemas (source and target) that need to be matched. It operates solely on schema-level information (attribute names, metadata) without access to raw data in the tables.  This is also a key difference compared to the table reasoning papers, which often rely on the actual data values for answering questions or verifying facts.

\textbf{Outputs:}
Table reasoning papers aim to output answers to questions, binary fact verification labels, updated tables after manipulation, generated SQL/code, etc. In contrast, Matchmaker outputs a mapping between the source and target schema attributes, or indicates no match is possible for certain attributes. The set of attribute pairs representing the schema matching results, can be used to guide data integration processes.

\textbf{Use of the LLM:}
Table reasoning employs LLMs for direct answer generation \citep{chen2023fewshot}, program synthesis \citep{nahid2024tabsqlify,kong2023opentab}, iterative prompting \citep{wang2023chaintable}, or as part of an agent system \citep{lu2024large}. Matchmaker uses LLMs for reasoning within a compositional program, generating candidates, refining them, and scoring confidence.

\textbf{Optimization/Training:}
Table reasoning works explore fine-tuning \citep{nahid2024tabsqlify}, instruction tuning \citep{lu2024large}, and in-context few-shot learning \citep{chen2023fewshot}. Matchmaker introduces a novel optimization process to select synthetic in-context examples for self-improvement without labeled data or fine-tuning.

\textbf{Key differences:}
In summary, while the table reasoning papers focus on tasks like question answering, fact verification, and table manipulation on single tables, Matchmaker addresses the distinct task of schema matching across table pairs. 
Its novel approach of a self-improving compositional language model program operating on schema-level information contrasts with general table reasoning which mostly use LLMs for direct table QA or program synthesis.

\newpage

\subsection{Matchmaker algorithm}\label{appendix:algorithm}

Below we provide a high-level overview algorithm of Matchmakers compositional language model program for schema matching.

\begin{algorithm}[h]
\caption{Matchmaker: Schema Matching with Self-Improving Compositional Language Model Programs}
\begin{algorithmic}[1]
\Require Source schema $S_s$, Target schema $S_t$
\Ensure Schema matches $M$
\State \textbf{Stage 1: Multi-Vector Document Creation}
\For{each table $T \in S_t$}
    \State Create document $D_T$ with attribute names and descriptions
    \State Append table metadata to $D_T$
    \State Encode $D_T$ using ColBERT-v2 to obtain multi-vector representation $V_T$
    \State Add $V_T$ to vector database $\mathcal{V}$
\EndFor
\State \textbf{Stage 2: Candidate Generation}
\For{each source attribute $q_i \in S_s$}
    \State Encode $q_i$ using ColBERT-v2 to obtain query embedding $E_{q_i}$
    \State Retrieve top-k semantic candidates $C_s$ from $\mathcal{V}$ using $E_{q_i}$
    \State Generate reasoning-based candidates $C_R$ using LLM $l_c(q_i, S_t)$
    \State Refine candidate set $C^* \leftarrow l_r(C_s \cup C_R, q_i)$
\EndFor
\State \textbf{Stage 3: Confidence Scoring}
\For{each source attribute $q_i \in S_s$}
    \State Format candidate set $C^*$ as multiple-choice question $Q_i$
    \For{each candidate $c_j \in C^*$}
        \State Compute confidence score $s_j \leftarrow l_s(Q_i, c_j)$
    \EndFor
    \State $m_i \leftarrow \operatorname{argmax} \limits_{c_j \in C^*} s_j$ \Comment{Select match with highest confidence}
    \State Add $(q_i, m_i)$ to schema matches $M$
\EndFor
\State \textbf{Stage 4: Self-Improvement Optimization}
\State Generate evaluation set $D_{\text{eval}}$ from unlabeled schemas
\For{each example $e_i \in D_{\text{eval}}$}
    \State $(\hat{y}_i, \text{trace}_i) \leftarrow \text{Matchmaker}(e_i)$ \Comment{Run Matchmaker to get output and traces}
    \State $s_i \leftarrow E_l(e_i, \hat{y}_i)$ \Comment{Compute evaluation score using LLM $E_l$}
    \State Add $(e_i, \text{trace}_i, \hat{y}_i, s_i)$ to $D_{\text{demo}}$
\EndFor
\State Sort $D_{\text{demo}}$ by score $s_i$
\State Select top-n examples from $D_{\text{demo}}$ as synthetic in-context examples
\State Update Matchmaker components with selected in-context examples
\State \Return Schema matches $M$
\end{algorithmic}
\label{alg:matchmaker}
\end{algorithm}

\newpage

\subsection{Schema matching challenges.}\label{appendix:challenges}

\begin{itemize}
\item \textbf{Database Heterogeneity}: The number of tables in each schema may differ, i.e., $|T_s| \neq |T_t|$, making it challenging to establish correspondences between attributes across schemas.
\item \textbf{Structural Heterogeneity}: Schemas may have different architectures, hierarchies, and representational granularity. If we define a hierarchy function $h(T_i)$ that describes the level of nesting within tables, differences in $h(T_{sj})$ and $h(T_{tk})$ for any $j$, $k$ can lead to significant challenges in aligning attributes $A_{sj}$ and $A_{tk}$.

\item \textbf{Semantic Heterogeneity}: Attributes in different schemas may have the same name but different meanings, or different names but the same meaning. Let $N_i = \{n_{ij} | A_{ij} \in A_i\}$ be the set of attribute names for schema $S_i$. Semantic heterogeneity occurs when $\exists A_{sj} \in A_s, A_{tk} \in A_t : f(A_{sj}) = A_{tk} \wedge n_{sj} \neq n_{tk}$ or when $\exists A_{sj} \in A_s, A_{tk} \in A_t : f(A_{sj}) \neq A_{tk} \wedge n_{sj} = n_{tk}$.

\item \textbf{Data Type Heterogeneity}: Attributes in different schemas may have different data types, even if they refer to the same concept. Let $d_{ij}$ be the data type of attribute $A_{ij}$. Data type heterogeneity occurs when $\exists A_{sj} \in A_s, A_{tk} \in A_t : f(A_{sj}) = A_{tk} \wedge d_{sj} \neq d_{tk}$.

\item \textbf{Information Mismatch}: Some attributes in one schema may lack a corresponding match in the other schema. This necessitates reasoning about "no possible match" cases, which is as important as reasoning about possible matches.

\item \textbf{Unsupervised Nature}: Schema matching is unsupervised, where no labeled data pairs $(A_{sj}, A_{tk})$ are available to train or validate the mappings. This necessitates reliance on the intrinsic structure and semantic information encoded in $A_i$, making the development of an effective mapping function $f$ challenging without external supervision.
\end{itemize}

\newpage
\subsection{Complexity of the MIMIC-OMOP task}

MIMIC-OMOP is a real-world healthcare schema matching task, which is reflective of complex structures, interlinking and hierarchies that can be expected in real-world schema matching tasks. Hence, Matchmakers ability to empirically outperform baselines on these tasks highlights its ability to handle complex schemas. 

\begin{figure}[!h]
    \vspace{-0mm}
    \centering
    \includegraphics[angle=90, width=0.9\linewidth]{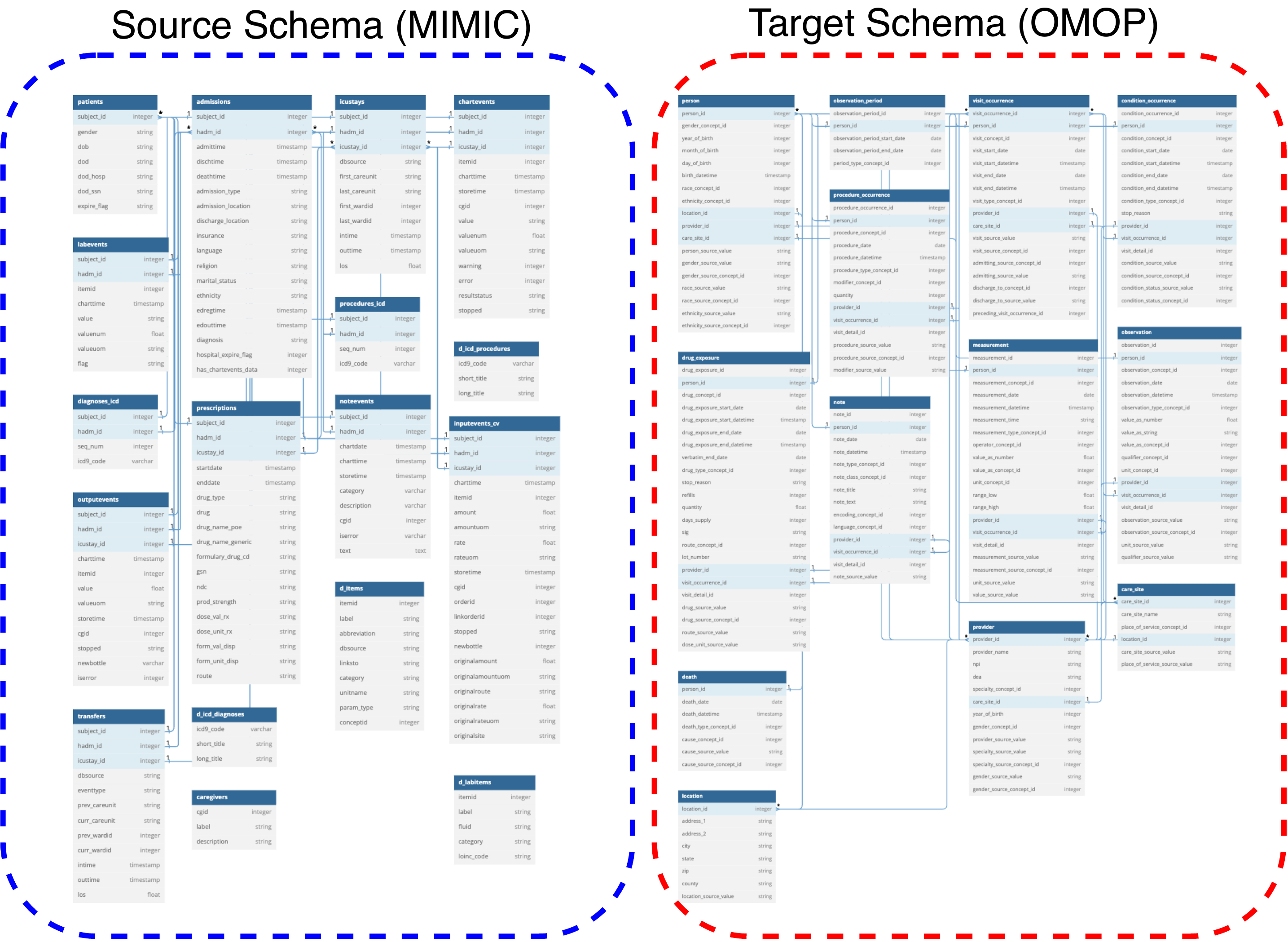}
    \caption{Illustration of the MIMIC-OMOP schema matching task showing the complexity and schema hierarchies.}
    \label{fig:mimic-omop-complexity}
    \vspace{-5mm}
\end{figure}

To illustrate the complexity of the schemas that Matchmaker can handle, Figure \ref{fig:mimic-omop-complexity} illustrates the complex schema structure and multiple tables.

\clearpage
\newpage

\section{Experimental details: Benchmarks \& datasets}\label{appendix:details}

All experiments are run on a single Nvidia A4000 GPU with 20 GB of vram. We invoke GPT-4 via the Azure OpenAI API.

\subsection{Benchmarks}

\subsubsection{Matchmaker}

Matchmaker is a compositional language model program for schema matching made up of multiple component modules --- formulated in the context of information retrieval. 

\textbf{GPT-4 Hyper-parameters.} The model version used as the LLM was GPT-4-1106, with the following settings:
\{'temperature': 0.5,
 'max\_tokens': 1024,
 'top\_p': 1,
 'frequency\_penalty': 0,
 'presence\_penalty': 0,
 'n': 1,
 \}

 \textbf{Embedding model and documents.} We use Colbert-V2 \citep{santhanam2022colbertv2} as the embedding model and follow the document creation process as outlined in Sec. \ref{rag}. We use the implementation of Colbert-v2 from RAGatouille (https://github.com/bclavie/RAGatouille/).

 \textbf{Candidates.} For both semantic and reasoning-based candidates, we set k=5.

\textbf{Optimization.} As described in the main paper, we generate synthetic in-context samples to address the unique challenges of a lack of labeled data and no demonstrations. As described, to achieve this we follow a boostrapping process like in DSPy \citep{khattab2023dspy}. For our experiments we select at maximum 4 synthetic in-context examples

\textbf{Prompts:} We show examples with the prompts for each component of Matchmaker in Appendix \ref{appendix:examples}.

\subsubsection{ReMatch}

In the main text we report the numbers directly from the ReMatch paper, as there is no open-source implementation. 

\textbf{How we selected the numbers to report:} The ReMatch paper does an exploration of the number of documents retrieved. Hence, we use the following two criteria.\\
(i) At least 1 document must be retrieved. i.e. the retrieval step cannot be skipped.

(ii) We then select the result that satisfies (i), with the highest accuracy@5. 

Our implementation of ReMatch follows the original paper \citep{sheetrit2024rematch}. We use OpenAI Ada embeddings for the embedding model and GPT-4 as the LLM.

We following the document creation procedure and use the prompt template as provided.

\textbf{GPT-4 Hyper-parameters.} The model version used for generation was GPT-4-1106, with the following settings from the ReMatch paper:
\{seed=42, temperature=0.5, max\_tokens=4096, top\_p=0.9, frequency\_penalty=0, presence\_penalty=0\}

\subsubsection{Jellyfish}

Jellyfish \citep{zhang2023jellyfish} is a fine-tuned language model tailored for data preprocessing tasks including schema matching. The 7B and 13B models are fine tuned upon the OpenOrca-Platypus2 model.

Implementation (7b): https://huggingface.co/NECOUDBFM/Jellyfish-7B

Implementation (13b): https://huggingface.co/NECOUDBFM/Jellyfish-13B

\subsubsection{LLM-DP}

LLM-DP \citep{Narayan2022CanFM, Zhang2023LargeDP} refer to works which have used pre-trained LLMs like GPT-3.5 or GPT-4 for data processing tasks like schema matching via prompting.  Since the papers in the few-shot case use labeled examples we do not use those --- given its unrealistic in practice. Hence, for these baselines they operate in a zero shot manner.

Implementation: https://github.com/HazyResearch/fm\_data\_tasks

\subsubsection{SMAT}

SMAT is a supervised learning approach which performs schema matching via an attention mechanism. Of course, the model needs labeled data to train on. In our experiments, we assess two variants given that labeled training data for schema matching is hard to access: (i) 20-80: 20\% train and 80\% test and (ii) 50-50: 50\% train and 50\% test.

We use the default hyper-parameters: \{Learning Rate: 0.8, Batch Size: 64,
Epochs: 30\}

Implementation: https://github.com/JZCS2018/SMAT

\subsection{Datasets}

We outline the two real-world schema matching benchmarks used in this paper --- MIMIC and Synthea. These datasets mapping different clinical/healthcare schemas were chosen as they are the standard datasets used in schema matching literature and consequently, used by prior works providing fair assessment. They are also considered the most reflective of real-world schema matching complexity and challenges. We note that the scarcity of complex and challenging real-world datasets, underscores the challenges in collecting and annotating real-world schema matching data. For instance, as noted in Sec 1, annotating MIMIC-OMOP alone required 500 hours from two medical experts.

Table \ref{tab:datasets} provides a summary of the table properties.

Note there is no specific train-test sets used as in supervised learning. As we perform the schema matching task in a zero-shot manner.

\begin{table}[!h]
    \centering
        \caption{Summary of the table properties of our two schema matching datasets.}
    \begin{tabular}{c|cc}
        Dataset & Source Tables & 
        Target Tables    \\ \hline \hline 
        MIMIC-OMOP & 26 & 14   \\
        SYNTHEA-OMOP & 12  &  21  \\
    \end{tabular}

    \label{tab:datasets}
\end{table}

\textbf{MIMIC Dataset:} The dataset contains a schema mapping between the 
MIMIC-III electronic health record (Source schema) \citep{johnson2016mimic}  and The Observational
Medical Outcomes Partnership Common Data Model (OMOP schema) (Target schema). 

This dataset is currently the largest publicly
available schema matching dataset \citep{sheetrit2024rematch} and is the cloest to a real-world schema matching use case, wherein a proprietary database created for a specific purpose (a source schema) is mapped to a given industry standard (a target schema) for further uses. In this case the proprietary database schema is MIMIC and the industry standard is the OMOP common data model.

\textit{Open-source data}:  https://github.com/meniData1/MIMIC\_2\_OMOP

\textbf{Synthea Dataset:}  The Synthea dataset is part of the OMAP benchmark \citep{Zhang2021SMATAA} and is a partial mapping of the Synthea \citep{walonoski2018synthea} (Source Schema) which is a synthetic healthcare dataset of a Massachusetts health records and attempts to map it to a subset of the OMOP CDM (Target Schema). The dataset has widely been used in previous schema matching papers \citep{sheetrit2024rematch, Narayan2022CanFM,Zhang2021SMATAA} as a realistic and challenging real-world schema matching benchmark. 

\textit{Open-source data}: https://github.com/JZCS2018/SMAT/tree/main/datasets/omap/

\clearpage
\newpage

\section{Examples using Matchmaker (with prompts)}\label{appendix:examples}
\vspace{-1mm}
\subsection{Matchmaker prompt examples}
\vspace{-1mm}
We show two end-to-end schema matching examples with Matchmaker, where other methods fail. (1) Example 1: case with No possible target schema match for the source schema query, (2) Example 2: challenging reasoning case, where there is a match possible between source and target schema.\\
$\blacktriangleright$ \textbf{In each component, we can show the "Optimized" In-context examples.}

\subsubsection{Example 1.}
\vspace{-1mm}
\textbf{Source schema query:} admissions-marital\_status(string): Table admissions details-the admissions table
gives information regarding a patient’s admission to the hospital., Attribute marital\_status details
-describe patient demographics. 

\textbf{Target scheme match:}  None possible.

\textbf{Matchmaker:}  None of the above.

\vspace{-2mm}
\begin{figure}[!h]
\caption{EXAMPLE 1: Candidate generation.}
\centering
\scalebox{0.9}{
\begin{promptbox}{Candidate generation}

You are an OMOP Schema expert. Your goal is to take the OMOP schema and based on the input, refine the schema to include only 5 most likely matches to the input query.

---

Follow the following format.

Input Schema: Input OMOP schema values
Input Query: input query
Refined Schema: Five most likely matches to input query. Include most likely matches to the input query. Respond with a single JSON object. JSON Schema: \{"defs": \{"Extractor": \{"properties": \{"related": \{"description": "related matches", "title": "Related", "type": "string"\}\}, "required": ["related"], "title": "Extractor", "type": "object"\}\}, "properties": \{"value": \{"items": \{"ref": "/defs/Extractor"\}, "title": "Value", "type": "array"\}\}, "required": ["value"], "title": "Output", "type": "object"\}

---

Input Schema: ['visit\_occurrence-person\_id(bigint)', 'visit\_occurrence-visit\_occurrence\_id(bigint)', ...  'procedure\_occurrence-provider\_id(bigint)', 'visit\_detail-visit\_detail\_source\_value(varchar(50))']

Input Query: procedureevents\_mv-itemid

Refined Schema: {"value": [{"related": "procedure\_occurrence-person\_id(bigint)"}, {"related": "procedure\_occurrence-visit\_occurrence\_id(bigint)"}, {"related": "procedure\_occurrence-procedure\_date(date)"}, {"related": "procedure\_occurrence-procedure\_source\_value(varchar(50))"}, {"related": "procedure\_occurrence-procedure\_concept\_id(integer)"}]}

---

Input Schema: ['visit\_occurrence-person\_id(bigint)', 'visit\_occurrence-visit\_occurrence\_id(bigint)', ...  'procedure\_occurrence-provider\_id(bigint)', 'visit\_detail-visit\_detail\_source\_value(varchar(50))']

Input Query: noteevents-text

Refined Schema: {"value": [{"related": "note-note\_text(varchar(max))"}, {"related": "note-note\_title(varchar(250))"}, {"related": "note-note\_source\_value(varchar(50))"}, {"related": "note-note\_date(date)"}, {"related": "note-note\_datetime(datetime)"}]}

---

Input Schema: ['visit\_occurrence-person\_id(bigint)', 'visit\_occurrence-visit\_occurrence\_id(bigint)', ...  'procedure\_occurrence-provider\_id(bigint)', 'visit\_detail-visit\_detail\_source\_value(varchar(50))']

Input Query: admissions-marital\_status(string): Table admissions details-the admissions table gives information regarding a patient’s admission to the hospital., Attribute marital\_status details -describe patient demographics.

Refined Schema:

-----------------------------------------------------------------

LLM Response

----------------------------------------------------------------

\{"value": [\{"related": "person-person\_id(bigint)"\}, \{"related": "person-gender\_source\_value(varchar(50))"\}, \{"related": "person-birth\_datetime(datetime)"\}, \{"related": "person-death\_datetime(datetime)"\}, \{"related": "visit\_occurrence-admitted\_from\_source\_value(varchar(50))"\}]\}

\end{promptbox}}

\end{figure}

\begin{figure}[!h]
\centering
\caption{EXAMPLE 1: Candidate refinement.}
\scalebox{0.85}{
\begin{promptbox}{Candidate refiner}
You are an expert OMOP matching ranker. Your task is to take the OMOP candidates and based on the input, refine the candidates to select the 5 most likely matches to the input query. Return ONLY the keys.

---

Follow the following format.

Input Schema: List of key: value pairs

Input Query: input query

Reasoning: Let's think step by step in order to produce the refined string list. We ...

Refined String List: Five most likely matches to input query. Include maximum of the 5 most likely matches to the input query. Return ONLY the keys

---

Input Schema: ['person-person\_id(bigint): Table person details-this table serves as the central identity management for all persons in the database. it contains records that uniquely identify each person or patient, and some demographic information., Attribute person\_id details -it is assumed that every person with a different unique identifier is in fact a different person and should be treated independently.', ...  'measurement-measurement\_concept\_id(integer): the measurement\_concept\_id field is recommended for primary use in analyses, and must be used for network studies']

Input Query: procedureevents\_mv-itemid

Reasoning: Let's think step by step in order to produce the refined string list. We are looking for a match to 'procedureevents\_mv-itemid', which seems to relate to a procedure event and its item identifier. The 'procedureevents\_mv' part suggests we're looking at a procedure occurrence, and the 'itemid' part suggests we're looking for an identifier or id field. Given that, we should look for keys in the 'procedure\_occurrence' table that represent identifiers.

Refined String List: 'procedure\_occurrence-person\_id(bigint)', 'procedure\_occurrence-visit\_occurrence\_id(bigint)', 'procedure\_occurrence-procedure\_concept\_id(integer)', 'procedure\_occurrence-visit\_detail\_id(bigint)', 'procedure\_occurrence-provider\_id (bigint)'

---

Input Schema: ['person-person\_id(bigint): Table person details-this table serves as the central identity management for all persons in the database. it contains records that uniquely identify each person or patient, and some demographic information., Attribute person\_id details -it is assumed that every person with a different unique identifier is in fact a different person and should be treated independently.', ...  'measurement-measurement\_concept\_id(integer): the measurement\_concept\_id field is recommended for primary use in analyses, and must be used for network studies']

Input Query: noteevents-text

Reasoning: Let's think step by step in order to produce the refined string list. We are looking for matches to "noteevents-text" in the input schema. The term "noteevents" matches with "note" in the input schema, and "text" matches with "note\_text". So, we should consider those keys that contain both "note" and "text".

Refined String List: ['note-note\_text(varchar(max))', 'note-note\_title(varchar(250))', 'note-note\_source\_value(varchar(50))', 'note-note\_text(varchar(max))']

---

Input Schema: ['person-person\_id(bigint): Table person details-this table serves as the central identity management for all persons in the database. it contains records that uniquely identify each person or patient, and some demographic information., Attribute person\_id details -it is assumed that every person with a different unique identifier is in fact a different person and should be treated independently.', ...  'measurement-measurement\_concept\_id(integer): the measurement\_concept\_id field is recommended for primary use in analyses, and must be used for network studies']

Input Query: admissions-marital\_status(string): Table admissions details-the admissions table gives information regarding a patient’s admission to the hospital., Attribute marital\_status details -describe patient demographics.

Reasoning: Let's think step by step in order to

-----------------------------------------------------------------

LLM Response

----------------------------------------------------------------

"produce the refined string list. We are looking for a match to 'admissions-marital\_status', which seems to relate to a hospital admission and the marital status of the patient. The 'admissions' part suggests we're looking at a visit occurrence or visit detail, and the 'marital\_status' part suggests we're looking for demographic information, which is usually found in the 'person' table. Given that, we should look for keys in the 'visit\_occurrence', 'visit\_detail', and 'person' tables.

Refined String List: 'visit\_occurrence-person\_id(bigint)', 'visit\_occurrence-visit\_occurrence\_id(bigint)', 'visit\_detail-person\_id(bigint)', 'visit\_detail-visit\_occurrence\_id(bigint)
\end{promptbox}}
 
\end{figure}

\vspace{-10mm}

\begin{figure}[!h]
\centering
\caption{EXAMPLE 1: MCQ Formatter.}
\scalebox{0.8}{
\begin{promptbox}{MCQ Formatter}
You are an expert MCQ formatter. Your task is to take a list of schema values and convert them into a multiple choice question format with (letter)Schema value, where the schema values should be key(description).

---

Follow the following format.

Input: input list of schema values
Mcq: MCQ format of schema values e.g (A)Schema value, (B)Schema value. Do not include additional options, only the schema values as options. where the schema values should be key(description). Add a No Match option.

---

Input: ["'procedure\_occurrence-person\_id(bigint)', 'procedure\_occurrence-visit\_occurrence\_id(bigint)', 'procedure\_occurrence-procedure\_concept\_id(integer)', 'procedure\_occurrence-visit\_detail\_id(bigint)', 'procedure\_occurrence-provider\_id (bigint)'"]
Mcq: (A)'procedure\_occurrence-person\_id(bigint)', (B)'procedure\_occurrence-visit\_occurrence\_id(bigint)', (C)'procedure\_occurrence-procedure\_concept\_id(integer)', (D)'procedure\_occurrence-visit\_detail\_id(bigint)', (E)'procedure\_occurrence-provider\_id (bigint)', (F)No Match

---

Input: ["['note-note\_text(varchar(max))', 'note-note\_title(varchar(250))', 'note-note\_source\_value(varchar(50))', 'note-note\_text(varchar(max))']"]
Mcq: (A)note-note\_text(varchar(max)), (B)note-note\_title(varchar(250)), (C)note-note\_source\_value(varchar(50)), (D)note-note\_text(varchar(max)), (E)No Match

---

Input: ["'visit\_occurrence-person\_id(bigint)', 'visit\_occurrence-visit\_occurrence\_id(bigint)', 'visit\_detail-person\_id(bigint)', 'visit\_detail-visit\_occurrence\_id(bigint"]
Mcq:

-----------------------------------------------------------------

LLM Response

----------------------------------------------------------------

"(A)'visit\_occurrence-person\_id(bigint)', (B)'visit\_occurrence-visit\_occurrence\_id(bigint)', (C)'visit\_detail-person\_id(bigint)', (D)'visit\_detail-visit\_occurrence\_id(bigint)', (E)No Match",
\end{promptbox}}

\end{figure}

\vspace{-10mm}

\begin{figure}[!h]
\centering
\caption{EXAMPLE 1: Confidence scoring.}
\scalebox{0.8}{
\begin{promptbox}{Confidence scoring}
You are a schema matching expert. Your task is given the input and the MCQ format of the schema, predict the likelihood or relation score from 0-100 of the input query being related to each option. Your scores will be calibrated. If there is no good match score No Match as 100

---

Follow the following format.

Input Mcq: Input MCQ format of schema values

Input Query: input query

Relation: Relation score of input query being related to the option as value. Assess each independently including No Match, returning a score from 0-100 for each. Return with key as MCQ letter e.g (A) and score=value as JSON

---

Input Mcq: (A)'procedure\_occurrence-person\_id(bigint)', (B)'procedure\_occurrence-visit\_occurrence\_id(bigint)', (C)'procedure\_occurrence-procedure\_concept\_id(integer)', (D)'procedure\_occurrence-visit\_detail\_id(bigint)', (E)'procedure\_occurrence-provider\_id (bigint)', (F)No Match

Input Query: procedureevents\_mv-itemid

Relation: \{"(A)": 0, "(B)": 0, "(C)": 0, "(D)": 0, "(E)": 0, "(F)": 100\}

---

Input Mcq: (A)note-note\_text(varchar(max)), (B)note-note\_title(varchar(250)), (C)note-note\_source\_value(varchar(50)), (D)note-note\_text(varchar(max)), (E)No Match

Input Query: noteevents-text

Relation: \{"A": 90, "B": 0, "C": 0, "D": 90, "E": 10\}

---

Input Mcq: (A)'visit\_occurrence-person\_id(bigint)', (B)'visit\_occurrence-visit\_occurrence\_id(bigint)', (C)'visit\_detail-person\_id(bigint)', (D)'visit\_detail-visit\_occurrence\_id(bigint', (E)No Match
Input Query: admissions-marital\_status(string): Table admissions details-the admissions table gives information regarding a patient’s admission to the hospital., Attribute marital\_status details -describe patient demographics.

Relation:

-----------------------------------------------------------------

LLM Response

----------------------------------------------------------------

"\{\"(A)\": 0, \"(B)\": 0, \"(C)\": 0, \"(D)\": 0, \"(E)\": 100\}"
\end{promptbox}}

\end{figure}

\clearpage
\newpage

\subsubsection{Example 2}

\textbf{Source schema query:} admissions-marital\_status(string): Table admissions details-the admissions table
gives information regarding a patient’s admission to the hospital., Attribute marital\_status details
-describe patient demographics. \\

\textbf{Target scheme match:} ’procedure\_occurrence-
quantity \\

\textbf{Matchmaker:}  ’procedure\_occurrence-
quantity

\begin{figure}[!h]
\caption{Candidate generation.}
\centering
\scalebox{0.9}{
\begin{promptbox}{EXAMPLE 2: Candidate generation.}
You are an OMOP Schema expert. Your goal is to take the OMOP schema and based on the input, refine the schema to include only 5 most likely matches to the input query.

---

Follow the following format.

Input Schema: Input OMOP schema values
Input Query: input query
Refined Schema: Five most likely matches to input query. Include maximum of the 10 most likely matches to the input query. Respond with a single JSON object. JSON Schema: \{"defs": \{"Extractor": \{"properties": \{"related": \{"description": "related matches", "title": "Related", "type": "string"\}\}, "required": ["related"], "title": "Extractor", "type": "object"\}\}, "properties": \{"value": \{"items": \{"ref": "/defs/Extractor"\}, "title": "Value", "type": "array"\}\}, "required": ["value"], "title": "Output", "type": "object"\}

---

Input Schema: ['visit\_occurrence-person\_id(bigint)', 'visit\_occurrence-visit\_occurrence\_id(bigint)', ....  'visit\_detail-visit\_detail\_source\_value(varchar(50))']
Input Query: procedureevents\_mv-itemid
Refined Schema: \{"value": [\{"related": "procedure\_occurrence-person\_id(bigint)"\}, \{"related": "procedure\_occurrence-visit\_occurrence\_id(bigint)"\}, \{"related": "procedure\_occurrence-procedure\_date(date)"\}, \{"related": "procedure\_occurrence-procedure\_source\_value(varchar(50))"\}, \{"related": "procedure\_occurrence-procedure\_concept\_id(integer)"\}]\}

---

Input Schema: ['visit\_occurrence-person\_id(bigint)', 'visit\_occurrence-visit\_occurrence\_id(bigint)', ....  'visit\_detail-visit\_detail\_source\_value(varchar(50))']
Input Query: noteevents-text
Refined Schema: \{"value": [\{"related": "note-note\_text(varchar(max))"\}, \{"related": "note-note\_title(varchar(250))"\}, \{"related": "note-note\_source\_value(varchar(50))"\}, \{"related": "note-note\_date(date)"\}, \{"related": "note-note\_datetime(datetime)"\}]\}

---

Input Schema: ['visit\_occurrence-person\_id(bigint)', 'visit\_occurrence-visit\_occurrence\_id(bigint)', ....  'visit\_detail-visit\_detail\_source\_value(varchar(50))']
Input Query: procedures\_icd-seq\_num(integer): Table procedures\_icd details-contains icd procedures for patients, most notably icd-9 procedures, Attribute seq\_num details -provides the order in which the procedures were performed
Refined Schema:

"\{\"value\": [\{\"related\": \"procedure\_occurrence-person\_id(bigint)\"\}, \{\"related\": \"procedure\_occurrence-visit\_occurrence\_id(bigint)\"\}, \{\"related\": \"procedure\_occurrence-procedure\_date(date)\"\}, \{\"related\": \"procedure\_occurrence-procedure\_concept\_id(integer)\"\}, \{\"related\": \"procedure\_occurrence-quantity(integer)\"\}]\}"

\end{promptbox}}

\end{figure}

\begin{figure}[!h]
\caption{EXAMPLE 2: Candidate Refinement.}
\centering
\scalebox{0.9}{
\begin{promptbox}{Candidate refinement}
You are an expert OMOP matching ranker. Your task is to take the OMOP candidates and based on the input, refine the candidates to select the 5 most likely matches to the input query. Return ONLY the keys.

---

Follow the following format.

Input Schema: List of key: value pairs

Input Query: input query

Reasoning: Let's think step by step in order to \{produce the refined\_string\_list\}. We ...

Refined String List: Five most likely matches to input query. Include maximum of the 5 most likely matches to the input query. Return ONLY the keys

---

Input Schema: ['procedure\_occurrence-person\_id(bigint): Table procedure\_occurrence details-this table contains records of activities or processes ordered by, or carried out by, a healthcare provider on the patient with a diagnostic or therapeutic purpose., Attribute person\_id details -the person\_id of the person for whom the procedure is recorded. this may be a system generated code.' .... 'visit\_detail-care\_site\_id(bigint): this field provides information about the care site where the visit detail took place']

Input Query: procedureevents\_mv-itemid

Reasoning: Let's think step by step in order to produce the refined string list. We are looking for a match to 'procedureevents\_mv-itemid', which seems to relate to a procedure event and its item identifier. The 'procedureevents\_mv' part suggests we're looking at a procedure occurrence, and the 'itemid' part suggests we're looking for an identifier or id field. Given that, we should look for keys in the 'procedure\_occurrence' table that represent identifiers.

Refined String List: 'procedure\_occurrence-person\_id(bigint)', 'procedure\_occurrence-visit\_occurrence\_id(bigint)', 'procedure\_occurrence-procedure\_concept\_id(integer)', 'procedure\_occurrence-visit\_detail\_id(bigint)', 'procedure\_occurrence-provider\_id (bigint)'

---

Input Schema: ['procedure\_occurrence-person\_id(bigint): Table procedure\_occurrence details-this table contains records of activities or processes ordered by, or carried out by, a healthcare provider on the patient with a diagnostic or therapeutic purpose., Attribute person\_id details -the person\_id of the person for whom the procedure is recorded. this may be a system generated code.' .... 'visit\_detail-care\_site\_id(bigint): this field provides information about the care site where the visit detail took place']

Input Query: noteevents-text

Reasoning: Let's think step by step in order to produce the refined string list. We are looking for matches to "noteevents-text" in the input schema. The term "noteevents" matches with "note" in the input schema, and "text" matches with "note\_text". So, we should consider those keys that contain both "note" and "text".

Refined String List: ['note-note\_text(varchar(max))', 'note-note\_title(varchar(250))', 'note-note\_source\_value(varchar(50))', 'note-note\_text(varchar(max))']

---

Input Schema: ['procedure\_occurrence-person\_id(bigint): Table procedure\_occurrence details-this table contains records of activities or processes ordered by, or carried out by, a healthcare provider on the patient with a diagnostic or therapeutic purpose., Attribute person\_id details -the person\_id of the person for whom the procedure is recorded. this may be a system generated code.' .... 'visit\_detail-care\_site\_id(bigint): this field provides information about the care site where the visit detail took place']

Input Query: procedures\_icd-seq\_num(integer): Table procedures\_icd details-contains icd procedures for patients, most notably icd-9 procedures, Attribute seq\_num details -provides the order in which the procedures were performed

Reasoning: Let's think step by step in order to

"produce the refined string list. We are looking for a match to 'procedures\_icd-seq\_num', which seems to relate to a procedure occurrence and its sequence number. The 'procedures\_icd' part suggests we're looking at a procedure occurrence, and the 'seq\_num' part suggests we're looking for a sequential number or order field. Given that, we should look for keys in the 'procedure\_occurrence' table that represent order or sequence.

 Refined String List: 'procedure\_occurrence-person\_id(bigint)', 'procedure\_occurrence-visit\_occurrence\_id(bigint)', 'procedure\_occurrence-procedure\_date(date)', 'procedure\_occurrence-procedure\_concept\_id(integer)', 'procedure\_occurrence-quantity(integer)'"
\end{promptbox}}

\end{figure}

\vspace{-5mm}
\begin{figure}[!h]
\caption{EXAMPLE 2: MCQ Formatter.}
\centering
\scalebox{0.85}{
\begin{promptbox}{MCQ Formatter}
You are an expert MCQ formatter. Your task is to take a list of schema values and convert them into a multiple choice question format with (letter)Schema value, where the schema values should be key(description).

---

Follow the following format.

Input: input list of schema values
Mcq: MCQ format of schema values e.g (A)Schema value, (B)Schema value. Do not include additional options, only the schema values as options. where the schema values should be key(description). Add a No Match option.

---

Input: ["'procedure\_occurrence-person\_id(bigint)', 'procedure\_occurrence-visit\_occurrence\_id(bigint)', 'procedure\_occurrence-procedure\_concept\_id(integer)', 'procedure\_occurrence-visit\_detail\_id(bigint)', 'procedure\_occurrence-provider\_id (bigint)'"]
Mcq: (A)'procedure\_occurrence-person\_id(bigint)', (B)'procedure\_occurrence-visit\_occurrence\_id(bigint)', (C)'procedure\_occurrence-procedure\_concept\_id(integer)', (D)'procedure\_occurrence-visit\_detail\_id(bigint)', (E)'procedure\_occurrence-provider\_id (bigint)', (F)No Match

---

Input: ["['note-note\_text(varchar(max))', 'note-note\_title(varchar(250))', 'note-note\_source\_value(varchar(50))', 'note-note\_text(varchar(max))']"]
Mcq: (A)note-note\_text(varchar(max)), (B)note-note\_title(varchar(250)), (C)note-note\_source\_value(varchar(50)), (D)note-note\_text(varchar(max)), (E)No Match

---

Input: ["'procedure\_occurrence-person\_id(bigint)', 'procedure\_occurrence-visit\_occurrence\_id(bigint)', 'procedure\_occurrence-procedure\_date(date)', 'procedure\_occurrence-procedure\_concept\_id(integer)', 'procedure\_occurrence-quantity(integer)'"]
Mcq:

"(A)'procedure\_occurrence-person\_id(bigint)', (B)'procedure\_occurrence-visit\_occurrence\_id(bigint)', (C)'procedure\_occurrence-procedure\_date(date)', (D)'procedure\_occurrence-procedure\_concept\_id(integer)', (E)'procedure\_occurrence-quantity(integer)', (F)No Match",
\end{promptbox}}

\end{figure}

\vspace{-10mm}
\begin{figure}[!h]
\caption{EXAMPLE 2: Confidence scoring.}
\centering
\scalebox{0.85}{
\begin{promptbox}{Confidence scoring}
You are a schema matching expert. Your task is given the input and the MCQ format of the schema, predict the likelihood or relation score from 0-100 of the input query being related to each option. Your scores will be calibrated. If there is no good match score No Match as 100

---

Follow the following format.

Input Mcq: Input MCQ format of schema values
Input Query: input query
Relation: Relation score of input query being related to the option as value. Assess each independently including No Match, returning a score from 0-100 for each. Return with key as MCQ letter e.g (A) and score=value as JSON

---

Input Mcq: (A)'procedure\_occurrence-person\_id(bigint)', (B)'procedure\_occurrence-visit\_occurrence\_id(bigint)', (C)'procedure\_occurrence-procedure\_concept\_id(integer)', (D)'procedure\_occurrence-visit\_detail\_id(bigint)', (E)'procedure\_occurrence-provider\_id (bigint)', (F)No Match
Input Query: procedureevents\_mv-itemid
Relation: \{"(A)": 0, "(B)": 0, "(C)": 0, "(D)": 0, "(E)": 0, "(F)": 100\}

---

Input Mcq: (A)note-note\_text(varchar(max)), (B)note-note\_title(varchar(250)), (C)note-note\_source\_value(varchar(50)), (D)note-note\_text(varchar(max)), (E)No Match
Input Query: noteevents-text
Relation: \{"A": 90, "B": 0, "C": 0, "D": 90, "E": 10\}

---

Input Mcq: (A)'procedure\_occurrence-person\_id(bigint)', (B)'procedure\_occurrence-visit\_occurrence\_id(bigint)', (C)'procedure\_occurrence-procedure\_date(date)', (D)'procedure\_occurrence-procedure\_concept\_id(integer)', (E)'procedure\_occurrence-quantity(integer)', (F)No Match
Input Query: procedures\_icd-seq\_num(integer): Table procedures\_icd details-contains icd procedures for patients, most notably icd-9 procedures, Attribute seq\_num details -provides the order in which the procedures were performed
Relation:

"\{\"(A)\": 0, \"(B)\": 0, \"(C)\": 0, \"(D)\": 0, \"(E)\": 90, \"(F)\": 10\}",
\end{promptbox}}

\end{figure}

\clearpage
\newpage

\subsection{LLM Evaluator}

We provide examples of the LLM evaluator, showing demonstrations achieving high and low scores.

\begin{figure}[!h]
\centering
\scalebox{0.9}{
\begin{evalbox}{LLM Evaluator Example: High Score}

You are a schema matching expert, your task is to rate if any of the suggested matches are potential good matches for the query. Be lenient and rate a match as good (4 or 5) if it is relevant to the query. Rate the matches from 1-5. If none of the matches are good, rate 0.

---

Follow the following format.

Query: The query.

Answers: possible matches

Reasoning: Let's think step by step in order to \{produce the rating\}. We ...

Rating: Rate if any of the suggested matches are good for the query from 1-5. Only output the rating and nothing else.

---

Query: noteevents-chartdate

Answers: ["'note-note\_date(date)', 'note-note\_datetime(datetime)', \\ 'note-visit\_occurrence\_id(bigint)', 'note-provider\_id(bigint)', 'note-person\_id(bigint)'"]

Reasoning: Let's think step by step in order to

-----------------------------------------------------------------

LLM Response

----------------------------------------------------------------

produce the rating. We can see that the query is looking for a match related to the date of a note event in a chart. The first two suggested matches, 'note-note\_date(date)' and 'note-note\_datetime(datetime)', are both related to dates and notes, making them potentially good matches for the query. The other three suggestions are not directly related to dates or note events, so they are not as good matches.\\
Rating: 4

\end{evalbox}}
\caption{LLM evaluator example, rated with a high score.}
    \label{fig:eval_high}
\end{figure}

\vspace{-1mm}
\begin{figure}[!h]
\centering
\scalebox{0.9}{
\begin{evalbox}{LLM Evaluator Example: Low Score}
You are a schema matching expert, your task is to rate if any of the suggested matches are potential good matches for the query. Be lenient and rate a match as good (4 or 5) if it is relevant to the query. Rate the matches from 1-5. If none of the matches are good, rate 0. 

---

Follow the following format.

Query: The query.

Answers: possible matches

Reasoning: Let's think step by step in order to \{produce the rating\}. We ...

Rating: Rate if any of the suggested matches are good for the query from 1-5. Only output the rating and nothing else.

---

Query: cptevents-subsectionheader

Answers: ['visit\_occurrence-visit\_type\_concept\_id(int)',  \\
'visit\_detail-visit\_detail\_concept\_id(int)', 'procedure\_occurrence-procedure\_concept\_id(int)',  'procedure\_occurrence-visit\_detail\_id(int)', 'condition\_occurrence-visit\_occurrence\_id(int)']

Reasoning: Let's think step by step in order to

-----------------------------------------------------------------

LLM Response

----------------------------------------------------------------

"produce the rating. We can see that the query is looking for a match related to 'cptevents-subsectionheader'. However, none of the suggested matches seem to directly relate to this query. They all refer to different concepts such as 'visit\_occurrence', 'visit\_detail', 'procedure\_occurrence', and 'condition\_occurrence'. Therefore, none of the matches seem to be a good fit for the query.\\
Rating: 0

\end{evalbox}}
\caption{LLM evaluator example, rated with a low score.}
    \label{fig:eval_low}
\end{figure}

\clearpage
\newpage

\section{Additional experiments} \label{appendix:more-exps}

\subsection{Number of LLM calls}\label{llm-calls}

\textbf{Goal.} To compare the number of LLM calls required by Matchmaker and other baseline methods for schema matching on the MIMIC-OMOP and SYNTHEA-OMOP datasets.

\textbf{Experiment.} We count the number of LLM calls made by each method during the schema matching process on both the MIMIC-OMOP and SYNTHEA-OMOP datasets. For methods that do not rely on LLMs (e.g., SMAT), we consider the number of forward passes through the neural network as equivalent to an LLM call for comparison purposes.

\textbf{Results.} Table \ref{tab:llm-calls} presents the number of LLM calls required by each method on the two datasets. 

\begin{table}[!h]
    \centering
    \caption{Number of LLM calls}
    \begin{tabular}{l|cc}
    Method     & MIMIC-OMOP & SYNTHEA-OMOP \\
        \hline \hline
    Matchmaker     & 1340 & 890 \\
    ReMatch     & 268 & 178 \\
    Jellyfish-13b     & 24771 & 29637 \\
    Jellyfish-7b      & 24771 & 29637 \\
    LLM-DP     & 24771 & 29637 \\
    SMAT     & 24771 & 29637\\
    \end{tabular}
    
    \label{tab:llm-calls}
\end{table}

\textbf{Discussion.} The results in Table \ref{tab:llm-calls} highlight the efficiency of Matchmaker and ReMatch in terms of the number of LLM calls required for schema matching. 

Both Matchmaker and ReMatch formulate schema matching as an information retrieval problem, which significantly reduces the search space compared to the binary classification formulation used by Jellyfish-13b, Jellyfish-7b, LLM-DP, and SMAT.

The high number of LLM calls required by Jellyfish-13b, Jellyfish-7b, LLM-DP, and SMAT can be attributed to their formulation of schema matching as a binary classification problem over the Cartesian product of source and target attributes. In this formulation, the LLM is prompted to provide a label of Yes/No for each pair of source-target attributes, resulting in a large number of LLM calls that scales ($O(n^2)$). Consequently, these methods are computationally expensive and less scalable compared to Matchmaker and ReMatch, which employ a more efficient approach.

The fewer number of LLM calls used by Matchmaker and ReMatch has practical implications in terms of computational cost and runtime efficiency. By reducing the number of LLM calls, these methods can perform schema matching more quickly and with lower computational overhead compared to methods that rely on a large number of calls. This is particularly important when dealing with large-scale schemas or when schema matching needs to be performed frequently in real-world applications.

\subsection{Matchmaker with other LLMs}

\textbf{Goal.} To understand the performance of Matchmaker when using a less powerful LLM backbone compared to GPT-4, and contrast it with the ReMatch baseline using GPT-4.

\textbf{Experiment.} We evaluate the performance of Matchmaker using GPT-3.5 as the backbone LLM for all components, instead of GPT-4 which was used in the main experiments. We compare this to the performance of Matchmaker with GPT-4 and ReMatch with GPT-4. All other aspects of the setup remain the same as in the main text.

\textbf{Results.}  Table \ref{tab:gpt3-res} shows the schema matching accuracy@k for the different methods. We observe that Matchmaker with GPT-3.5 performs worse than Matchmaker with GPT-4, which is expected given GPT-3.5 is a less powerful LLM. Interestingly, Matchmaker with GPT-3.5 achieves comparable performance to ReMatch with GPT-4, despite GPT-3.5 being a much weaker LLM than GPT-4. On MIMIC, Matchmaker with GPT-3.5 slightly outperforms ReMatch with GPT-4 for accuracy@1 and is competitive at higher k. On Synthea, performance is similar for accuracy@1 but Matchmaker with GPT-3.5 outperforms ReMatch with GPT-4 for accuracy@3 and accuracy@5.

\begin{table}[!h]
    \vspace{-5mm}
    \centering
        \caption{\footnotesize Comparison of schema matching performance of different baselines.}
        \scalebox{0.75}{
    \begin{tabular}{c|c|cc|c}
        & & Matchmaker (GPT-4) & Matchmaker (GPT-3.5) & ReMatch (GPT-4)  \\
        \hline \hline
        \multirow{3}{*}{\rotatebox{90}{MIMIC}} & acc@1 &  \bf 62.20 $\pm$  2.40 \textcolor{ForestGreen}{$\uparrow$} & 48.30$\pm$ 2.80 \textcolor{ForestGreen}{$\uparrow$} & 42.50    \\
        & acc@3 & \bf  68.80 $\pm$ 2.00 & 62.00 $\pm$ 4.20 & 63.80     \\
        & acc@5 &  \bf  71.10 $\pm$ 2.00 & 70.00 $\pm$ 4.20 & \bf  72.90  \\
         \hline\hline
        \multirow{3}{*}{\rotatebox{90}{Synthea}} & acc@1 & \bf  70.20 $\pm$ 1.70  & 47.80 $\pm$   3.20 & 50.50     \\
        & acc@3 & \bf  78.60 $\pm$ 2.50 & 63.30 $\pm$ 4.30 \textcolor{ForestGreen}{$\uparrow$} & 58.10  \\
        & acc@5 & \bf  80.90 $\pm$ 1.10 & 77.10 $\pm$ 0.70 \textcolor{ForestGreen}{$\uparrow$} & 74.30   \\
    \end{tabular}}

    \label{tab:gpt3-res}
    \vspace{-2mm}
\end{table}

\textbf{Discussion.} These results demonstrate that the Matchmaker approach of using a compositional LLM program is quite robust and can provide good schema matching performance even with weaker LLM backbones. The fact that Matchmaker with GPT-3.5 is competitive with ReMatch using GPT-4 highlights the strength of the multi-stage Matchmaker approach over ReMatch's single-stage LLM usage. However, using a more powerful LLM like GPT-4 still provides significant gains, underlining the importance of using an LLM with powerful reasoning capabilities for this complex task.

\subsection{Further performance results: ReMatch reimplementation}

\textbf{Goal.} To compare the performance of Matchmaker against the ReMatch baseline, using both the original reported results from the ReMatch paper and the re-implementation of ReMatch.

\textbf{Experiment.} In the main paper, we report the performance of the ReMatch baseline using the results directly from the paper, as code is not available for ReMatch. However, for completeness, we also re-implement the ReMatch approach based on the details provided in the ReMatch paper.

Our re-implementation uses the OpenAI Ada-002 text embeddings for the retrieval step, following the same procedure as ReMatch for chunking and creating documents. We then use the same prompts as described in the ReMatch paper for the schema matching task. We compare the performance of our re-implemented ReMatch with the original reported results and Matchmaker.

\textbf{Results.} Table \ref{tab:more-results} presents the schema matching accuracy@k for Matchmaker, the original ReMatch results, and our re-implemented ReMatch. We observe that Matchmaker consistently outperforms both the original ReMatch results and our re-implementation across all metrics and datasets. We also find the re-implemented ReMatch achieves lower performance compared to the original reported results.

\begin{table}[!h]
    \vspace{-5mm}
    \centering
        \caption{\footnotesize Comparison of schema matching performance of different baselines.}
        \scalebox{0.75}{
    \begin{tabular}{c|c|c|cc}
        & & Matchmaker & ReMatch (Original) & ReMatch (Reimplemented)  \\
        \hline \hline
        \multirow{3}{*}{\rotatebox{90}{MIMIC}} & acc@1 &  \bf 62.20 $\pm$  2.40 & 42.50  & 41.99 $\pm$ 0.61  \\
        & acc@3 & \bf  68.80 $\pm$ 2.00 & 63.80  & 46.63 $\pm$ 1.99  \\
        & acc@5 &  \bf  71.10 $\pm$ 2.00 & \bf  72.90  & 46.63 $\pm$ 1.99 \\
         \hline\hline
        \multirow{3}{*}{\rotatebox{90}{Synthea}} & acc@1 & \bf  70.20 $\pm$ 1.70  & 50.50  & 29.10 $\pm$ 0.80   \\
        & acc@3 & \bf  78.60 $\pm$ 2.50 & 58.10  & 32.71 $\pm$ 0.35 \\
        & acc@5 & \bf  80.90 $\pm$ 1.10 & 74.30  & 33.46  $\pm$ 0.63 \\
    \end{tabular}}
    \label{tab:more-results}
    \vspace{-2mm}
\end{table}

\textbf{Discussion.} These results further confirm the superiority of Matchmaker over the ReMatch baseline, even when considering our re-implementation of the method. The lower performance of the re-implemented ReMatch compared to the original reported results could be due to differences in implementation details, such as the choice of text embeddings or variations not accounted for.
However, it is important to note that even with these differences, Matchmaker consistently outperforms ReMatch (original) by a significant margin. 
The fact that Matchmaker achieves strong performance gains over both the original ReMatch and our re-implementation underscores the value of the novel techniques introduced in Matchmaker, such as the multi-stage language model program, the use of diverse candidate generators and the self-improvement mechanism through synthetic in-context examples.

\subsection{Improving performance: Use of Existing Mappings to remedy errors}

\textbf{Goal.} To investigate the potential performance improvement in Matchmaker when leveraging readily available mappings to rectify errors between directly mapped attributes.

\textbf{Experiment.} In schema matching, certain attributes like source\_value and concept\_id have a direct mapping (e.g. in OMOP). If Matchmaker incorrectly maps the source attribute to the wrong target attribute (e.g., mapping to source\_value instead of concept\_id or vice versa), these errors can be easily rectified by leveraging the existing relationship.

To simulate this error correction, we implement a post-processing step where we adjust Matchmaker's predictions if the predicted target attribute has a direct mapping to the true target attribute. We apply this correction for all values of k and measure the resulting performance improvement.

\begin{figure}[!h]
    \centering
        \includegraphics[width=0.7\textwidth]{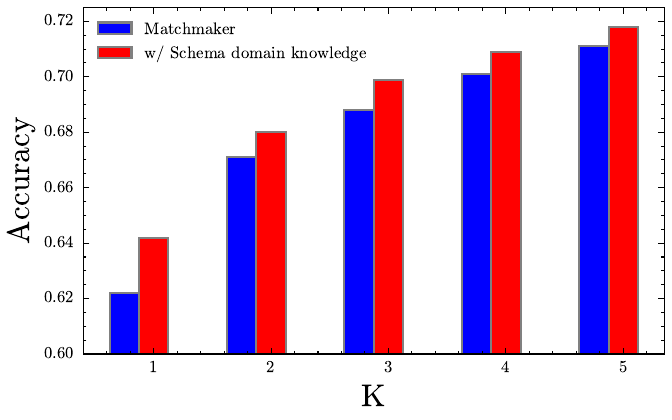}
    \caption{Performance improvement in Matchmaker when leveraging readily available mappings to correct errors between directly mapped attributes like source\_value and concept\_id.
}
    \label{fig:ontology}
\end{figure}

\textbf{Results.} Figure \ref{fig:ontology} shows the accuracy gains across different values of k when applying the mapping correction. We observe consistent performance improvements across all k values. These results indicate that leveraging knowledge can indeed help rectify some of the errors made by Matchmaker.

\textbf{Discussion.} While the results demonstrate the potential benefit of using existing mappings for error correction, it is important to note that the performance gains are relatively modest compared to other strategies like human-in-the-loop deferral based on Matchmaker's confidence scores (as shown in the main text).

Moreover, the mapping correction relies on the availability of direct mappings between attributes, which may not always exist in practice. Therefore, while this approach can serve as a useful post-processing step, it should be seen as a complementary technique to be used alongside other strategies like human-in-the-loop for improving schema matching performance.

\subsection{Comparison of Matchmaker on ontology matching tasks}

While Schema matching and ontology matching are seemingly related, in reality they are completely different tasks. Specifically, schema and ontology matching fundamentally differ in their task and available information. Ontology matching leverages richer contextual info, including properties, axioms, rules, concept hierarchies and additional annotations. In contrast, schemas are sparser, with only attribute names, data types, descriptions and links.

Despite the difference for completeness we evaluate recent LLM ontology match methods using GPT-4 backbones to mirror Matchmaker namely:  OLaLa \citep{hertling2023olala} and LLMs4OM \citep{giglou2024llms4om}.

As shown in Table \ref{onto-match}, Matchmaker outperforms these methods on both datasets.

\begin{table}[!h]
\vspace{-0mm}
\centering
\caption{Accuracy@1: Matchmaker vs two LLM-based Ontology matching methods.}
\scalebox{0.95}{
\begin{tabular}{l|c|c}
\hline
\textbf{Method} & \textbf{MIMIC} & \textbf{Synthea} \\ \hline
Olala & $33.58 \pm 0.47$ & $31.53 \pm 3.37$ \\ \hline
LLMs4OM & $44.78 \pm 0.41$ & $64.50 \pm 2.02$ \\ \hline
Matchmaker (Ours) &  $\bf 62.20 \pm \bf 2.40$ &  $\bf 70.20 \pm \bf 1.70$ \\ \hline
\end{tabular}
}
\label{onto-match}
\end{table}

\clearpage
\newpage

\section{Broader Impact}

Schema matching is a critical step in data integration, enabling the creation of large, harmonized datasets that can be used to train machine learning models. The proposed Matchmaker approach, with its self-improving compositional language model program, has the potential to significantly accelerate and automate the schema matching process, thus facilitating the development of more accurate and robust ML models across various domains.

The importance and value of schema matching cannot be overstated, as integrating data from various sources such as different regions, organizations or applications is vital in many fields, including healthcare, finance, and e-commerce. By enabling the integration of data from disparate sources, schema matching plays a critical role in creating comprehensive, harmonized datasets that can provide a more complete picture of the domain under study. For example, in healthcare, integrating data from multiple hospitals can lead to more representative and diverse datasets, allowing researchers to identify patterns and insights that may not be apparent when analyzing data from a single institution.

Moreover, schema matching is not only valuable for specific domains but also for the machine learning community as a whole. By increasing the pool of available data (larger and more diverse) for training and validation, schema matching can contribute to the development of more accurate, robust, and generalizable ML models.  Furthermore, having access to a larger pool of data can enable more rigorous validation and testing of ML models, allowing researchers to assess their performance across different subpopulations, time periods, and data sources. This, in turn, can lead to more reliable and trustworthy ML models that can be confidently applied in real-world settings.

Below we describe some positive implications that could be unlocked as schema matching approaches such as Matchmaker are used in practice. We also show some drawbacks with mitigation strategies.

\textbf{Positive Impacts:}
\begin{itemize}

\item Improved data integration: Matchmaker can help overcome the challenges of integrating data from heterogeneous sources, leading to the creation of larger, more comprehensive datasets. This can enable the development of more powerful and generalizable ML models.

\item  Accelerated research and discovery: By reducing the time and effort required for data integration, Matchmaker can accelerate research and discovery in fields, where data often resides in disparate databases with diverse schemas.

\item Enhanced decision-making: The ability to train ML models on larger, more diverse datasets enabled by Matchmaker can lead to more accurate and reliable predictions, supporting better decision-making in various applications.

\end{itemize}

\textbf{Potential Drawbacks and Mitigation Strategies:}
\begin{itemize}

\item Overreliance on automated schema matching: While Matchmaker can significantly automate the schema matching process, it is not perfect and may make errors. Overreliance on automated methods without human oversight could lead to incorrect data integration. Mitigation: Matchmaker should be used as a tool to assist human experts in the schema matching process, rather than as a complete replacement. The paper demonstrates how Matchmaker can be effectively used with a human-in-the-loop approach, leveraging the strengths of both human expertise and automated methods.

\item Propagation of errors: If Matchmaker introduces errors during the schema matching process, these errors can propagate downstream and affect the quality of the resulting integrated datasets and ML models. Mitigation: It is essential to implement rigorous validation and quality control measures to detect and correct errors introduced by Matchmaker. This can include manual spot-checks, automated consistency checks, and the use of domain-specific validation rules. Establishing a feedback loop to continuously monitor and improve Matchmaker's performance based on real-world usage can also help mitigate the propagation of errors.

\end{itemize}

\clearpage

\section*{NeurIPS Paper Checklist}
\begin{enumerate}
\item {\bf Claims}
    \item[] Question: Do the main claims made in the abstract and introduction accurately reflect the paper's contributions and scope?
    \item[] Answer: \answerYes{} 
    \item[] Justification: The abstract accurately reflects the claims made in the paper. Our paper introduces
    Matchmaker a language model program for schema matching which we introduce in detail in Sec.4. We then experimentally show in Sec. 5 on real-world and widely used schema matching datasets how Matchmaker compares to other alternatives. We also assess different components of Matchmaker, as well as, showing how it could be integrated with humans in practice. Overall, we believe this matches the claims. 
    \item[] Guidelines:
    \begin{itemize}
        \item The abstract and/or introduction should clearly state the claims made, including the contributions made in the paper and important assumptions and limitations. A No or NA answer to this question will not be perceived well by the reviewers. 
        \item The claims made should match theoretical and experimental results, and reflect how much the results can be expected to generalize to other settings. 
        \item It is fine to include aspirational goals as motivation as long as it is clear that these goals are not attained by the paper. 
    \end{itemize}

\item {\bf Limitations}
    \item[] Question: Does the paper discuss the limitations of the work performed by the authors?
    \item[] Answer: \answerYes{} 
    \item[] Justification: We include a discussion of limitations in Section 6. 

\item {\bf Theory Assumptions and Proofs}
    \item[] Question: For each theoretical result, does the paper provide the full set of assumptions and a complete (and correct) proof?
    \item[] Answer: \answerNA{} 
    \item[] Justification: We do not include explicit theoretical results or proofs. However, all mathematical formalism and equations in Section 3 and Section 4 are accompanied by underlying assumptions and rationale.

    \item {\bf Experimental Result Reproducibility}
    \item[] Question: Does the paper fully disclose all the information needed to reproduce the main experimental results of the paper to the extent that it affects the main claims and/or conclusions of the paper (regardless of whether the code and data are provided or not)?
    \item[] Answer: \answerYes{} 
    \item[] Justification: Experimental details are provided in Section 5, with further details in Appendix B. We also provide prompts and examples in Appendix C. The implementation of our method closely follows Section 4 and the algorithm in Appendix A.

\item {\bf Open access to data and code}
    \item[] Question: Does the paper provide open access to the data and code, with sufficient instructions to faithfully reproduce the main experimental results, as described in supplemental material?
    \item[] Answer: \answerNo{} 
    \item[] Justification: Besides the descriptions in Sec 5, we also provide details about the algorithms and data in Appendix B. 

\item {\bf Experimental Setting/Details}
    \item[] Question: Does the paper specify all the training and test details (e.g., data splits, hyperparameters, how they were chosen, type of optimizer, etc.) necessary to understand the results?
    \item[] Answer: \answerYes{} 
    \item[] Justification: All the details on data, hyper-parameters etc for the experiments are provided in Appendix B.

\item {\bf Experiment Statistical Significance}
    \item[] Question: Does the paper report error bars suitably and correctly defined or other appropriate information about the statistical significance of the experiments?
    \item[] Answer: \answerYes{} 
    \item[] Justification: Error bars (standard deviation) are included as relevant over multiple seeds for the experiments in Section 5 and
Appendix C. Stochasticity is due to the LLM temperature.

\item {\bf Experiments Compute Resources}
    \item[] Question: For each experiment, does the paper provide sufficient information on the computer resources (type of compute workers, memory, time of execution) needed to reproduce the experiments?
    \item[] Answer: \answerYes{} 
    \item[] Justification: The compute details on the experiments are provided in Appendix B. The number of LLM calls are detailed in Appendix C.
    
\item {\bf Code Of Ethics}
    \item[] Question: Does the research conducted in the paper conform, in every respect, with the NeurIPS Code of Ethics \url{https://neurips.cc/public/EthicsGuidelines}?
    \item[] Answer: \answerYes{} 
    \item[] Justification: We have read the code of  ethics do not violate any of the dimensions.

\item {\bf Broader Impacts}
    \item[] Question: Does the paper discuss both potential positive societal impacts and negative societal impacts of the work performed?
    \item[] Answer: \answerYes{} 
    \item[] Justification: We highlight broader impacts in Section 1 and 6 of the paper, as well as, having a dedicated broader impact statement in Appendix E. 
    
\item {\bf Safeguards}
    \item[] Question: Does the paper describe safeguards that have been put in place for responsible release of data or models that have a high risk for misuse (e.g., pretrained language models, image generators, or scraped datasets)?
    \item[] Answer: \answerNA{} 
    \item[] Justification:  Not applicable — our paper presents a new method for schema matching which doesn't have such risks.

\item {\bf Licenses for existing assets}
    \item[] Question: Are the creators or original owners of assets (e.g., code, data, models), used in the paper, properly credited and are the license and terms of use explicitly mentioned and properly respected?
    \item[] Answer: \answerYes{} 
    \item[] Justification: Appendix B provides details and/or citations for all assets (data and baselines) used in the paper.

\item {\bf New Assets}
    \item[] Question: Are new assets introduced in the paper well documented and is the documentation provided alongside the assets?
    \item[] Answer: \answerNA{} 
    \item[] Justification: The paper does not produce new assets such as datasets, but uses existing datasets/benchmarks.

\item {\bf Crowdsourcing and Research with Human Subjects}
    \item[] Question: For crowdsourcing experiments and research with human subjects, does the paper include the full text of instructions given to participants and screenshots, if applicable, as well as details about compensation (if any)? 
    \item[] Answer: \answerNA{} 
    \item[] Justification: We do not have crowdsourcing experiments or research with humans.

\item {\bf Institutional Review Board (IRB) Approvals or Equivalent for Research with Human Subjects}
    \item[] Question: Does the paper describe potential risks incurred by study participants, whether such risks were disclosed to the subjects, and whether Institutional Review Board (IRB) approvals (or an equivalent approval/review based on the requirements of your country or institution) were obtained?
     \item[] Answer: \answerNA{} 
    \item[] Justification: We do not have crowdsourcing experiments or research with humans that would need an IRB.

\end{enumerate}

\end{document}